# Exploring the relationship between human-centric AI and firm idiosyncratic risks

**Dr. Zhen-yuan (Ralph) Liu**

Department of Accounting, School of Economics and Management,

China University of Mining and Technology, Tongshan District, Xuzhou City, China

Email: 6635@cumt.edu.cn

**Ms. Yu-ting Wang**

School of Economics and Management,

Nanjing Forestry University, Xuanwu District, Nanjing City, China

Email: wyt0807@njfu.edu.cn

**Prof. Jia-jia Yan**

School of Economics and Management,

Fuzhou University, Minhou District, Fuzhou City, China

Email: 53896165@qq.com

**Prof. Shivam Gupta** *

Department of Information Systems, Supply Chain Management & Decision Support,

NEOMA Business School, Reims, France

Email: shivam.gupta@neoma-bs.fr

(* : Corresponding Author)

**Prof. Mihalis Giannakis**

Audencia Business School, Nantes, France

Email: mgiannakis@audencia.com

# Exploring the relationship between human-centric AI and firm idiosyncratic risks

**Abstract**: Despite the extensive discussions of human-centric AI (HCAI) in Industry 5.0, its effects on firms' idiosyncratic risks (IR) remains underexplored. This is an imperative issue for firms navigate financial risks during the current technological revolution, as IR reflects investor reactions to corporate heterogeneous AI strategies and implementations by isolating firm-level stock volatility from systematic factors. Integrating situated AI theory with social-technical systems theory, we conceptualise HCAI as a situated AI strategy that reduces AI-related ethical risks and fosters AI-Human synergies in firms' business operations, ultimately reducing IR by aligning with stakeholders' diverse expectations. Moreover, socio-technical factors, namely digitalisation, operational efficiency, executive shareholding, and CEOs with IT background, may moderate the HCAI-IR relationship. Using a multi-source panel dataset of Chinese listed firms from 2015 to 2023, we find that HCAI is associated with lower firm IR. Furthermore, digitalisation and executive shareholding strengthen this risk-reducing effect, whereas operational efficiency and CEOs with IT background surprisingly attenuate it. Our findings offer theoretical contributions and practical insights for both ethical AI governance and firm financial risk management in the AI era.

## 1. **Introduction**

A wide range of businesses have adopted Artificial Intelligence (AI) in recent years, with many firms using it to support decision-making and enhance daily operations. AI is a broad term with many possible definitions depending on context, and it refers to giving machines the capacity to operate in domains that demand human judgment, e.g., tasks such as problem-solving, decision-making, and creative generation (Hudson & Morgan, 2024; Madan & Ashok, 2025). Building on prior computer technologies, AI shows an unprecedented capacity to optimise operational processes, enhance strategic decision-making, and even drive business model innovations (Czarnitzki et al., 2023). Just as consumers have quickly embraced the possibilities of conversational AI in their daily lives, managers and employees have also incorporated AI into their corporate frameworks (Sun & Song, 2025).

However, the transformative potential of AI is still tempered by significant challenges, from generalisability, ethical issues and embedded bias to a tendency toward myopic optimisation. These issues can cause AI use to have significant repercussions (Vesnic-Alujevic et al., 2020), and together expose firms to financial, social, and ethical perils (Berente et al., 2021). The Commonwealth Bank of Australia (CBA) typifies the dangers involved, as the firm attempted to lay off 45 employees due to the integration of a new AI voice-bot system in July 2025, but reversed its decision a month later after admitting its initial assessments did not fully consider the business implications. The abortive layoffs led to decreased customer satisfaction, reputational damage, and increased operational costs[1]. Conversely, when IBM announced an AI partnership prominently featured with governance protocols, explainability standards, and human-in-the-loop design, its stock price rose sharply[2]. These contrast cases suggest that investors reward firms' AI strategies framed with human-centric orientation rather than workforce substitutions.

In response to growing concerns over AI risks, a new synergistic Human-Centric AI

---

[1] Source available at: https://www.hcamag.com/au/specialisation/hr-technology/cba-reverses-ai-triggered-employee-layoffs/546887

[2] Source available at: https://www.morningstar.com/news/marketwatch/2025100762/ibms-stock-rises-toward-a-record-why-its-anthropic-deal-symbolizes-a-new-frontier-in-ai

(HCAI) concept was defined as part of *Industry 5.0*, a proposed next phase of industrial development by the European Union in 2021. HCAI aims to enrich human-AI collaboration, moving the conversation from what is technically possible to ensuring responsible AI use (Marioni et al., 2024). The key objective reframes human-AI interaction, positioning AI as complement rather than replacement. This shift minimises the typical risks of using AI for business management and industrial systems, shoring up trust and ensuring a base level of suitability (Heyder et al., 2023). HCAI enables organisations to navigate complex internal structures and volatile market environments, while mitigating algorithmic bias and enhancing organisational resilience (Eitel-Porter, 2021; Heyder et al., 2023). In addition, by emphasising transparency, explainability, and ethical accountability (Bankins et al., 2024; Marioni et al., 2024), organisations adopting HCAI principles are more likely to generate social value through improved efficiency and fairness in decision-making and operational processes (Passalacqua et al., 2025). The guiding principle is that AI-driven performance gains should not compromise fairness or social responsibility. In this way, HCAI has moved from a purely conceptual model to an increasingly crucial strategic framework, enabling firms to secure a sustainable competitive edge in a fast-moving, disruptive digital era.

Despite most literature affirming the merits of HCAI across fields such as intelligent manufacturing optimisation (Xu et al., 2021), operational performance (Passalacqua et al., 2025), and sustainability (Liu et al., 2025), there is little research examining whether and how investors respond to corporate HCAI initiatives. Addressing this research gap is crucial, as investor assessments significantly affect firm decision-making that aims to balance emerging technology development and industrial transformation (Xu et al., 2020). Investor reactions can also hinder a firm's access to financial resources and exacerbate risks, thereby further obstructing its long-term prosperity. Reflecting investor responses to a firm’s future cash flows through stock price volatility, idiosyncratic risk (IR) accounts for over 80% of overall risk in the capital market. Further, IR is substantially distinct from systematic risk, assessing market-wide factors that affect all stocks, as it not only links more closely to specific firm-level factors but also serves as a robust measure of how investors react to firms’ strategic decisions (Moon et al., 2023). Therefore, IR has been extensively used as the predictor of firm risks across the fields of information systems (Berente et al., 2021), accounting and finance (Campbell et al., 2014; Herskovic et al., 2016; Li et al., 2014), economics (Asness et al., 2020; Hasan & Habib, 2017), and management studies (Nguyen et al., 2020). For instance, Li et al. (2021) unveil that CSR practices exhibit the U-shaped impact on firms' IR, the relationship of which could be flattened by firms' AI innovations. Hudson and Morgan (2024) reveal that firms' AI exposure significantly lowers IR, as both investors and financial analysts perceive AI as the core driving force behind future business, thereby bringing firms stable future cash flows. However, to the best of our knowledge, no prior study sheds light on the effect of HCAI on firms' IR during the Industry 5.0 transition. This gap is crucial, given that HCAI innovation and implementation programmes require substantial resources, capital and organisational change, potentially exposing firms to high execution risks. Further, more recent research emphasises the importance of identifying contingent factors that curb IR from corporate characteristics and internal governance structures (Aldawsari et al., 2025; Morgeson et al., 2024) and then enriches managerial and policy implications for managing firms' IR (He et al., 2025; Wang et al., 2024). To sum up, this study is guided by two research questions as follows:

***RQ 1: Whether and how does Human-centric AI affect firms' idiosyncratic risks?***

***RQ 2: If so, what are the contingent factors moderating the HCAI-IR relationship?***

To answer these questions, we integrate situated AI theory (SAIT) with the socio-technical system (STS) theoretical framework to view HCAI as firms' situated AI strategy during the Industry 5.0 transition. SAIT argues that only by overcoming the generic nature and myopia of AI, and by addressing the needs of broader stakeholders, can AI strategies provide firms with a competitive advantage (Kemp, 2024). In this vein, HCAI enhances employee skills, unlocks human-AI ensemble power, and fosters positive stakeholder relationships through human-centred design, ethical management, and responsible applications (Heyder et al., 2023). Investors observe these AI choices through disclosures and innovation outputs. Attention shifts toward long-term value, and firm-specific volatility (idiosyncratic risk) tends to fall. Meanwhile, SAIT further indicates that organisational constraints significantly moderate the effects of firms' AI strategies at both technological and socio levels. Therefore, we follow Bednar and Welch (2020) to explore contingent factors that potentially moderate the HCAI-IR relationship from both socio-technical perspectives. Specifically, we argue that more substantial risk reduction occurs when firms possess competitive advantages in both socio (i.e., executive shareholdings and CEO's IT background) and technical (i.e., digital transformation and operational efficiency) systems.

To testify these propositions, we follow Liu et al. (2025)'s work to identify HCAI innovations from 2.873 million patent textual data of Chinese listed firms, as HCAI remains at its early stage and the relevant technological innovations reflect firms' AI strategies more accurately (Miric et al., 2023). After econometrically analysing panel data containing 16,461 observations of Chinese listed firms from 2015 to 2023, we found that HCAI significantly reduces firms' IR. This finding remains robust after several examinations, including alternative measures, time lagging model, propensity score matching (PSM), and sub-sample regression methods. More importantly, the moderating analysis results further demonstrate that the risk-curbing effect of HCAI is amplified when firms possess higher levels of both digital transformation and executive shareholding. However, operational efficiency and IT background CEO weaken the reducing impact of HCAI on IR. One plausible explanation pertains to shareholders' pressure for throughput and a technology-first orientation of companies, which dilutes human oversight and cross-functional checks, leading investors to perceive higher execution risk.

Regarding the contributions of this study, we first contribute to the ongoing discussion around HCAI by unveiling its curbing effect on firms' IR. Prior work stresses the internal gains of HCAI in production systems, employee skill-upgrading, and sustainable performance. We extend this view and provide a novel perspective on how investors reflect upon and evaluate corporate HCAI innovations. Our evidence links design-led governance to this financial outcome and answers the special issue's call for work that connects AI design to financial risk analysis. Secondly, we identify the boundary conditions of the risk-reducing effects of HCAI related to technical and social dimensions. Digitalisation and executive shareholding strengthen the negative association between HCAI and IR, while operational efficiency and an IT background CEO weaken it. These results give practical guidance for HCAI implementation. Finally, this paper expands the organisational constraints of SAIT by integrating the socio-technical perspective and by anchoring claims in a finance construct observed by investors, thereby offering a theoretical implication for future research in this field.

## 2. Literature review and theoretical background

### 2.1 Human-centric AI (HCAI) in Industry 5.0 transition

AI played a crucial role in achieving Industry 4.0 and led to exponential gains in production efficiency through automation technologies (Mathew et al., 2023). When integrating AI into digital and smart transformation, firms obtain substantial operational advantages through automating routine tasks (Wamba-Taguimdje et al., 2020), boosting product innovation processes (Chowdhury et al., 2024; Jarrahi et al., 2023), and optimising resource allocations (Li et al., 2023; Yao et al., 2025). However, AI also exposes essential risks to firms due to its generality, ethical concerns, functional myopia (Berente et al., 2021), and lack of algorithmic transparency (Marcus, 2018). For instance, AI-based recruitment tools trained on historical data may systematically exclude certain groups (e.g., female technical candidates), as evidenced by the failure of Amazon's AI recruitment system (Dastin, 2022), which reveals fundamental conflicts of AI's automation-based logic that overlook human values (Cath et al., 2018).

The European Union launched its Industry 5.0 initiative to cope with these sorts of AI challenges, seeking to move humans back to the driver's seat in terms of decision-making and advocating for the use of "human-AI ensembles" (Koeppen et al., 2025; Nahavandi, 2019). HCAI serves as a core enabler here, with a more human-centred approach to AI technological development, innovation, and applications that address AI's inherent technical and societal limitations (Passalacqua et al., 2022). Specifically, HCAI is intended to assist in routine operations, freeing employees from repetitive tasks to focus on more complex work that demands greater situational awareness, decision oversight, or human-AI collaboration. This shift has the potential to revitalise the value and demand for such transformed roles. Core HCAI mechanisms also facilitate the capture, transformation, and utilisation of tacit knowledge that might otherwise be lost, thereby ensuring more sustainable organisational assets (Kemp, 2024; Xu et al., 2023). Unlike Industry 4.0, this approach specifically emphasises algorithmic transparency and explainability, alongside the embedding of values like fairness and safety that might otherwise be undervalued by an AI agent (Do Khac et al., 2025). This is exemplified by frameworks like the EU's "Trustworthy AI" (Jobin et al., 2019), which aims to rebuild trust that has been eroded by early human-AI interaction.

Beyond these broader HCAI concepts, more recent studies have specifically investigated the merits of HCAI in a management context. For instance, Psarommatis et al. (2023) propose the integration of human-centric principles into AI-enhanced Zero-Defect Manufacturing systems. Matthews et al. (2025) found that HCAI aligns firms' resources desire with broader stakeholder groups through AI synergies along the value chain. At a more micro level, HCAI could grant employees with effective human-AI collaborations, which boosts corporate social performance and sustainable business operations (Chowdhury et al., 2024). Liu et al. (2025) empirically examined the effect of HCAI on both social and operational performances of firms. However, to the best of our knowledge, the extant literature has not investigated whether and how investors react to firms' HCAI. Addressing this gap is crucial for managing firms' financial risks, particularly given that HCAI requires substantial resources, capital and organisational structures from investors. Accordingly, the primary objective of our study is to explore in depth the relationship between HCAI and investor reaction, the latter of which is commonly proxied by firm idiosyncratic risk in the extant literature.

### 2.2 Managing idiosyncratic risk (IR)

IR reflects investors' evaluation of and reaction to a firm's future cash flows via its stock price volatility (Bansal & Clelland, 2004). Distinct from systematic risk assessing market-wide factors that affect all firms, IR is tied more closely to firms' specific characteristics and accounts for more than 80% total risks in the capital market (Goyal & Santa-Clara, 2003). Also, IR serves as a robust indicator for assessing the efficacy of firms' strategic decisions (Moon et al., 2023), influencing firms' future cash flows and market values in either a favourable or a negative manner. Hence, effectively managing IR can safeguard investor and shareholder returns (Srinivasan & Hanssens, 2009; Xu & Malkiel, 2003), ensure stable cash flows, and mitigate the risk of bankruptcy (Lin & Shen, 2015). Previous research has explored a wide array of IR antecedents, ranging from IT strategy (Berente et al., 2021) and top management team characteristics (Bernile et al., 2018; Tan & Liu, 2016) to information disclosure (He et al., 2025; Miric et al., 2023; Moon et al., 2023; Reber et al., 2022) and CSR practices (Wang et al., 2024).

More recently, a burgeoning body of work has examined the link between AI and IR following three distinct perspectives in terms of AI exposure, AI design, and AI governance. In particular, studies adopting the exposure perspective treat AI primarily as the exogenous technological shock affecting firms' business operations. Hudson and Morgan (2024) find that greater AI exposure reduces firm IR, as financial analysts attribute technological changes to systematic rather than idiosyncratic risk. Conversely, Boido and Aliano (2025) argue that AI exposure may increase IR via heightening business operations uncertainty, given that adopting an AI-driven mindset consumes substantial resources and requires structural changes. While these exposure-oriented research offers valuable macro-level insights, they pay limited attention to how firms could strategically shape their AI deployments in ways that align with their characteristics and value-generating processes. Addressing this gap, Li et al. (2021) introduced the AI governance perspective, showing that AI innovations can flatten the U-shaped relationship between firms' CSR practices and IR. This finding highlights that the risk implications of AI are contingent upon how firms oversee, incentivise, and align AI with the interests of broader stakeholders. Building upon this pragmatic view, a nascent steam of research has begun to explore the design perspective, focusing on the intrinsic characteristics of AI systems that collectively constitute trustworthy and human-centric AI (Passalacqua et al., 2025).

However, to the best of our knowledge, no prior research has systematically and empirically investigated the effect of HCAI on firms' IR by integrating these three perspectives. Moreover, Morgeson et al. (2024) and Aldawsari et al. (2025) indicate that one of the main reasons for the current fragmented findings in the AI-IR relationship pertains to the omission of contingent factors rooted in corporate characteristics and internal governance structures. They advocate for more fine-grained investigations that identify when and under what conditions AI curbs IR, as such knowledge offers richer managerial and policy implications in AI era. In this regard, the second objective of our study is to explore contingent factors that may moderate the HCAI-IR relationship.

### 2.3 Integrating Situated AI theory with the socio-technical systems framework

Situated AI theory (SAIT; Kemp, 2024) provides us with an essential theoretical lens to accomplish the main research objectives above. Moving beyond the mere technology adoption perspective, SAIT focuses on how organisations can use AI strategy to acquire multi-fold competitive advantages. These strategies operate using a three-stage process: grounding, bounding, and recasting. Grounding AI refers to addressing its generic nature by deeply integrating it into a firm's operational processes, thereby enhancing resource allocation efficiency. Bounding AI seeks to undermine the possibility of AI performing unethical actions, such as data breaches, cybersecurity issues, and other unlawful encroachment. Finally, recasting AI is focused on reconfiguring stakeholder relationships to align the interests of broader stakeholders, thereby overcoming AI myopia.

Following these theoretical guidelines, HCAI can be understood as a corporate AI strategy aligned with Industry 5.0 requirements, which can be further delineated into three core constructs: human-centred design, ethical management, and responsible implications. Human-centred design prioritises human needs in AI development via creating user-friendly interaction interfaces as well as integrating ergonomic considerations. For instance, Covariant developed AI-driven robotic arm for collaborating with workers to process logistics parcel sorting and waste classification tasks in extreme environments[3]. Ethical management establishes base-level guardrails to address the challenges posed by AI's inscrutable nature, including algorithmic black boxes, biases, and privacy violations, thereby cultivating the trust necessary for effective human-AI collaboration. Simultaneously, responsible implications are pivoting towards a broader perspective in AI development, integrating stakeholder interests to counteract short-sighted decisions driven purely by efficiency gains.

---

[3] Source available at: https://covariant.ai/resources/introducing-the-next-phase-of-our-ai-robotics-journey

It is worth distinguishing HCAI from other established AI frameworks such as "trustworthy AI" or "responsible AI". Although the latter framework patterns articulate the ethical principles AI systems should satisfy, they remain largely principle-based and context-agnostic (Jobin et al., 2019). On the contrary, HCAI is firms' internal situating approach, as HCAI embeds those principles into firms' unique operational processes, governance mechanisms, and stakeholder relationship management, thereby answering how ethics can be operationalised through firms' proactive design, ethical issue management, and deployment choices during the Industry 5.0 transition. Figure 1 illustrates how we develop the core concepts of HCAI via the SAIT lens.

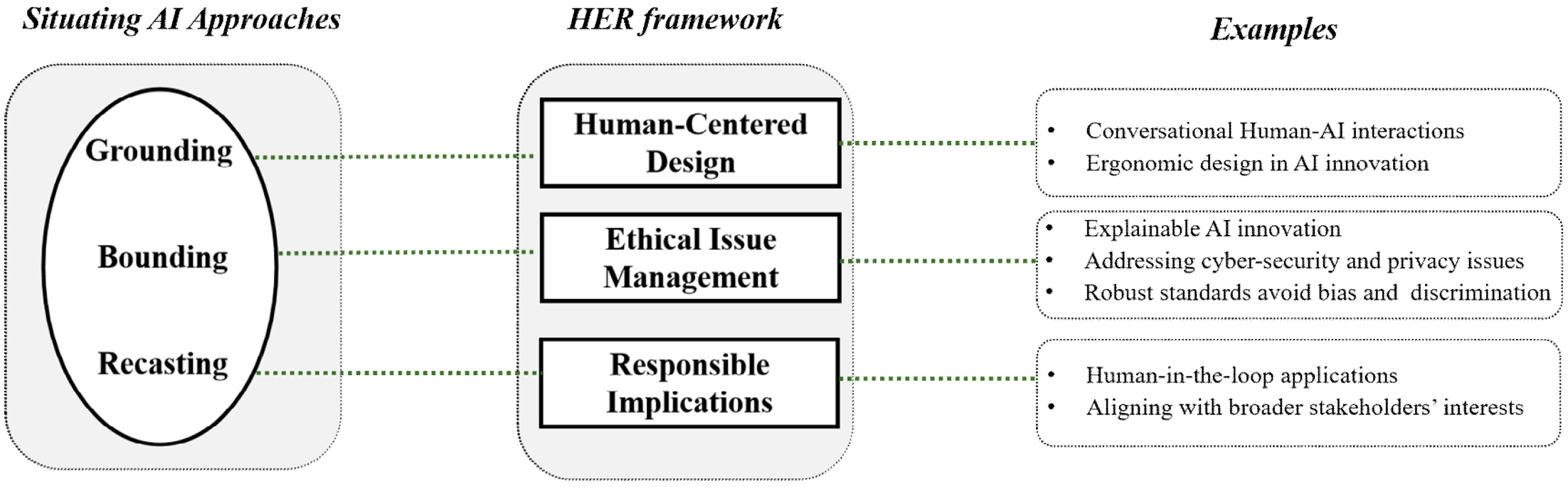

**Figure 1.** Leveraging SAIT to decipher HCAI

SAIT also indicates that organisational constraints moderate the effect of situated AI strategies. More importantly, Moser et al. (2024) pinpoint that SAIT needs to take the alignment between social and technical factors into account, especially when forming the comprehensive AI strategy for balancing stakeholders' expectations, which resonates profoundly with socio-technical systems (STS) theory. STS theory holds that only through the congruence between technical and socio subsystems can firms achieve superior performance (Dai et al., 2024). The technical subsystem comprises technology and task factors, while the socio subsystem involves organisational structures and people (Tong et al., 2023).

To operationalize STS theory in our moderation analysis, we screened one proxy from each of these four dimensions, ensuring balanced coverage of the potential boundary conditions that shape investor reactions to firms' HCAI. Specifically, we capture the technology dimension through digitalisation, which reflects the foundational digital infrastructure wherein HCAI is implemented. The task dimension is represented by firms' operational efficiency, as it reflects firms' task-level capability via measuring how systematically and effectively firms can converts inputs into pursued outputs. On the socio subsystems, we capture the structure dimension through executive shareholding, which aligns managerial incentives with long-term value creation and risk mitigation. Finally, the people dimension is represented by the CEO's IT background, which shapes leadership cognition and strategic orientations regarding technology deployment. Taken together, these four moderators allow us to comprehensively examine how the alignments within a firm's socio-technical contexts influence HCAI-IR relationship. Figure 2 presents the conceptual research model in this study.

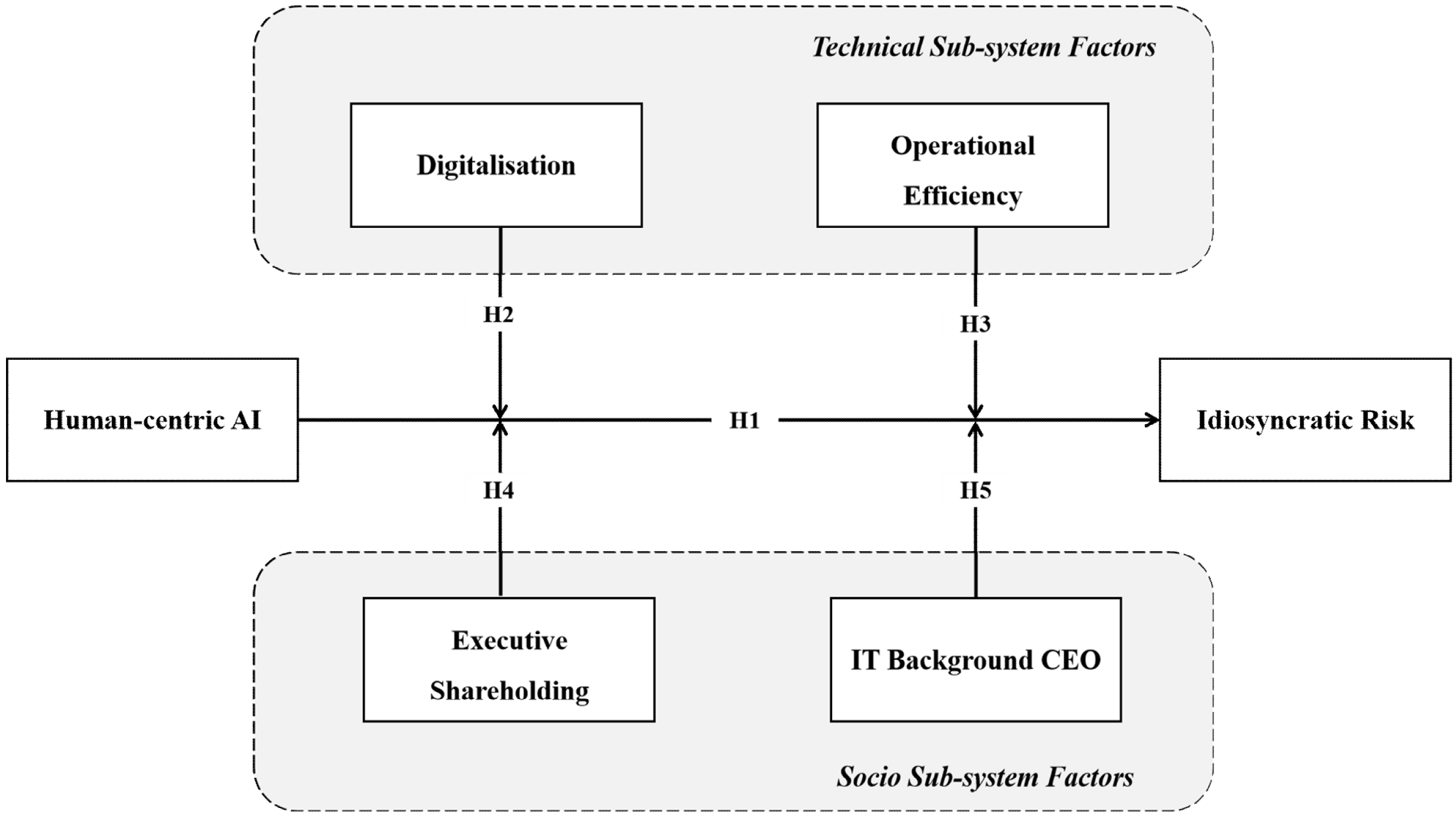

**Figure 2.** Conceptual research model

## 3. Hypothesis development

### 3.1 The effect of HCAI on IR

Drawing upon SAIT, we argue that HCAI could decrease firms' IR through at least three ways. First, investors particularly value HCAI when delving into target firms' AI strategy. Given the growing importance of sustainable investment, investors frequently rely on firms' sustainable performance for investment decisions (Eccles et al., 2020; Lee et al., 2022). When adopting HCAI, investors are aware of firms' efforts in balancing sustainable development requirements and AI technological pushes, improving the value of HCAI in hampering noise trading behaviour. Even though HCAI may impose additional processing costs on understanding firms' motivations, the human-AI collaborative methodologies, data sources, and weighting systems could bring firms' long-term returns (Wang et al., 2023). As a result, investors may judge a firm's actual value in HCAI, thereby decreasing its idiosyncratic risk.

Second, HCAI incorporates development approaches that help align a firm's AI ambitions with the interests of internal stakeholders, such as human-centred design and ethical management. This is intended to ensure sustainable growth and minimise future cash flow concerns. Specifically, human-centred design aims to augment employees with AI, rather than replacing them entirely (Matthews et al., 2025). Unsurprisingly, this attitude tends to enhance employee impressions of job security and even fosters innovation (Jia et al., 2024). Moreover, HCAI supports upskilling and reskilling initiatives, enabling employees to adapt to their evolving work demands in the AI era. This, in turn, mitigates operational risks associated with large-scale layoffs. For example, Haier, a large manufacturing firm in China, adopted HCAI principles in 2020 to transition employees into "autonomous entrepreneurs". This transformation increased employee innovation participation by 61% and reduced staff turnover by 40%. Such HCAI-driven innovation further enabled the company to achieve a growth rate five times the industry average between 2021 and 2024[4].

Third, HCAI has the potential to create synergistic benefits that extend beyond individual firms (Matthews et al., 2025) when understood from a multi-stakeholder viewpoint. In this way,

[4] Source available at:
https://ex.chinadaily.com.cn/exchange/partners/82/rss/channel/cn/columns/sz8srm/stories/WS67ee2468a310e29a7c4a78a1.html

value creation is a mutually reinforcing process where the success of each participant depends on the contributions of others (Bridoux & Stoelhorst, 2022). This interdependence is clearly reflected in HCAI principles, where intelligent systems require human collaboration and continuous data input to achieve optimal performance. The HCAI synergy effect therefore enhances value for broader stakeholders, while simultaneously increasing customer satisfaction and employee welfare, often without detriment to other parties (Tantalo & Priem, 2016). Moreover, adopting these principles may enhance coordination between upstream and downstream supply chain partners, thereby generating social value across the vertical business network. As AI investments can produce positive externalities, HCAI may gradually spread throughout the supply chain, contributing to improvements in collective social performance. Taken together, HCAI can serve as an important driver of more sustainable development paradigms. In this context, IR may be reduced when the focal firm adopts HCAI. Based on the above reasoning, we propose the following hypothesis:

***H1: Human-centric AI decreases firm idiosyncratic risk.***

### 3.2 The moderating factors

Despite the situating AI implementations for achieving competitive advantages, HCAI also needs to align with the social and technical subsystems a firm occupies. STS theory provides four components that should be analysed together when evaluating the organisational performance that emerging technologies could bring, i.e., technological foundation, operational processes, roles of people, and organisational structure (Colombari et al., 2024). Based on the extant studies on IR antecedents (Cao et al., 2024; Gu et al., 2019; Tan & Liu, 2016), we propose to investigate the contingent factors that potentially moderate the HCAI-IR relationship under the STS theoretical framework, which includes organisational digitalisation, operational efficiency, executive shareholding and CEO with IT background.

Digitalisation refers to the extent to which a firm undertakes technological digital transformation. Specifically, it signifies the deep integration of digital technologies with business processes, driving business model innovation and organisational change (Ancillai et al., 2023). Logically, firms with high levels of digitalisation possess a better foundation for integrating HCAI principles into their operational management and practices, as adopting HCAI enhances a range of digital technologies, to say nothing of its potential positive effects on the digital mindsets of employees. Therefore, we propose the following hypothesis:

***H2: A firm's digitalisation level moderates the relationship between HCAI and IR by amplifying the risk-reducing effect.***

Operational efficiency (OE) measures a firm's operational capability in terms of how it transforms resource inputs into expected outputs (Qin et al., 2025). Therefore, OE is an imperative indicator of long-term financial performance, as firms with higher OE are able to pursue organisational strategies and activities more efficiently and flexibly. Firms with higer OE are therefore better equipped to take full advantage of HCAI strategies, with the potential to optimise their operational processes using AI while minimising the traditional risks. This approach also delivers positive signals to investors and analysts in the capital markets, reinforcing the positive effect of HCAI on IR. However, efficient firms often operate with lean buffers and tightly coupled routines, leaving little slack resources for unforeseen adjustments. This may potentially induce investors to perceive HCAI in a operations-disrupting rather than risk-reducing manner. Therefore, we propose the third hypothesis as follows:

***H3: A firm's operational efficiency moderates the relationship between HCAI and IR via amplifying the risk-curbing effect.***

Executive shareholding refers to the extent to which a firms' shares are held by C-level executives as well as directors and board supervisors (Lyu et al., 2025), attenuating the myopic behaviors of managers and motivating managers to implement digital technologies in more ethical ways (Deng et al., 2025). In this regard, HCAI aligns with top managers' expectations when they hold more shares in firms and focus on sustainable development, which mitigates the risk-countering effects on firms' IR. Thereby, we propose that:

***H4: The more top managers hold a firm's share, the stronger risk-curbing effect HCAI generate on IR.***

CEOs' previous personal and professional experiences and knowledge fundamentally influence a firm's subsequent strategic orientations and related risks (Bendig et al., 2023; Wang et al., 2020). When a firm's CEO has an IT background, she may lay more emphasis on digital technologies, R&D investment and technology-driven operations. This professional background could navigate CEOs to obtain more profound insights regarding HCAI benefits, strengthening the risk-reducing effect of HCAI on IR. Interestingly, CEOs with backgrounds in IT are more likely to trust technological approaches like HCAI to solve obstacles facing the Industry 5.0 transition (Zhang et al., 2025). Yet, the technology-centric leadership style may also introduce countervailing dynamics. CEOs with strong IT backgrounds may instinctively prioritize automation, algorithmic optimization, and efficiency gains over the human-centric values that distinguish HCAI from generic AI. This also has a knock-on effect on how well a leadership team can understand HCAI, thereby allowing more informed decisions that eventually result in more traditional positive indicators such as profits and firm growth. Following this logic, we propose:

***H5: IT background CEOs moderate the relationship between HCAI and IR through enhancing the risk-reducing effect.***

## 4. Methodology

### 4.1 Data and samples

As a rapidly developing economic power, China provides a unique opportunity to empirically test our hypotheses. The Chinese government launched a strategic push to integrate AI into intelligent manufacturing in 2015, aligning with the national "Made in China 2025" initiative (Hao, 2024). Under this policy, China experienced exponential growth in AI innovation. One way to visualise this is depicted in Figure 3, which shows that the number of AI patents held by Chinese listed companies increased by nearly an order of magnitude over an eight-year span, from 676 in 2015 to 6,577 in 2023. One year later, President Xi announced China's Human-Centric AI Initiative, which mirrors the human-centred principles of the EU's *Industry 5.0* and the UN's *Governing AI for Humanity* agenda. While AI adoption in China continues at pace, specific characteristics of Chinese stock market may pose challenges to adopting HCAI principles, such as its high proportion of retail investors, prevalent speculative trading, and insider trading issues (He et al., 2025). These unique institutional factors have the potential to influence how effectively HCAI principles align with the transition to Industry 5.0. With both these headwinds and tailwinds in mind, these unique conditions make the study of Chinese listed firms a compelling sample for testing the hypotheses above.

In addition, the Chinese capital market is characterised by relatively stringent disclosure requirements for technological innovation activities, including patent filings, digital transformation initiatives, and corporate governance practices. Listed firms are required to provide detailed disclosures in annual reports and regulatory filings, which enhances the visibility of AI-related strategies to investors. This institutional environment facilitates investors' assessment of firms' human-centric AI efforts and strengthens the informational link between HCAI innovation and market-based risk perceptions.

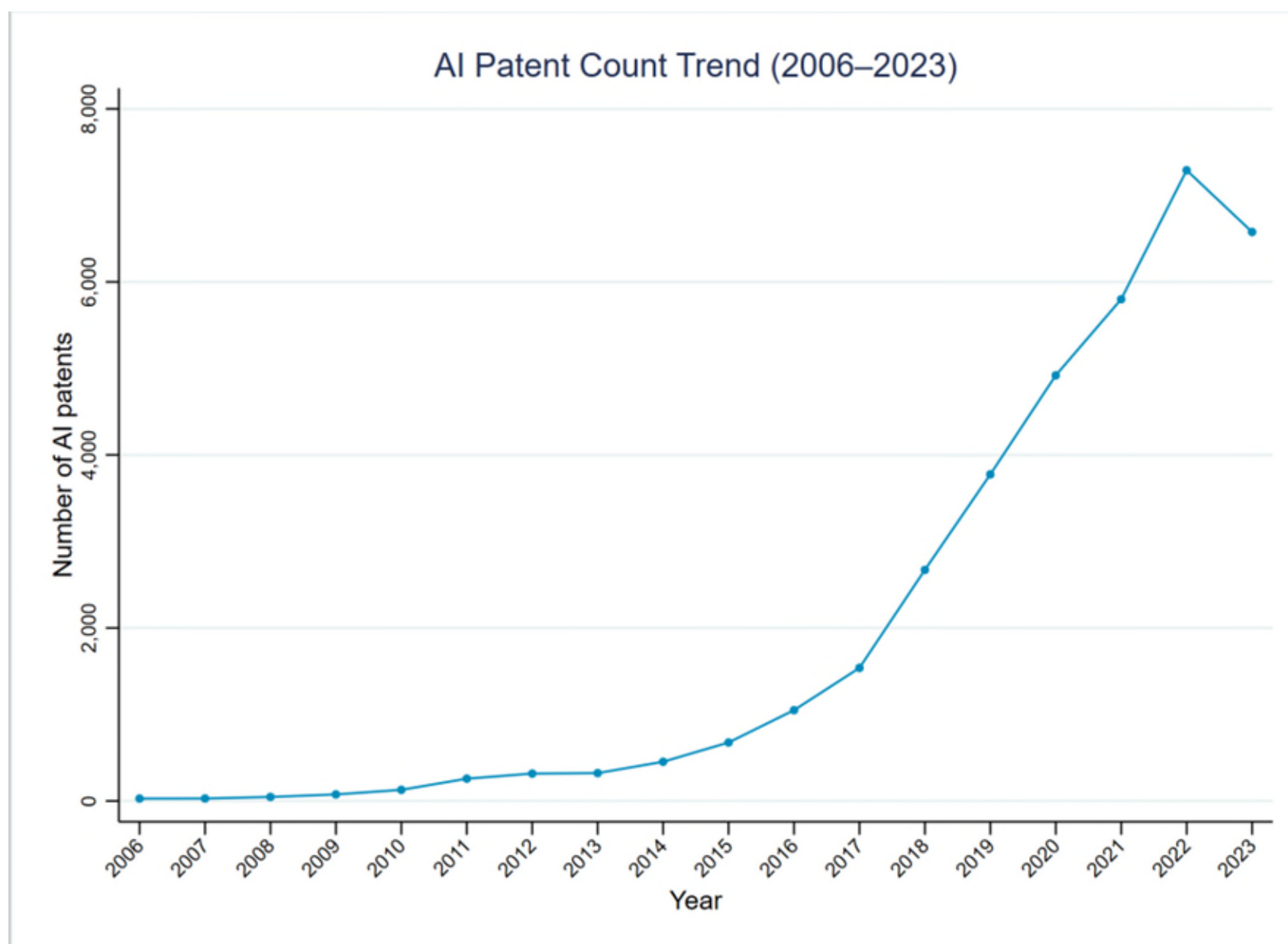

**Figure 3.** AI innovation Patent in China

Our sample includes all firms listed on the Shanghai and Shenzhen A-share markets between 2015 and 2023. These firms were evaluated in terms of HCAI innovation based on their patent records from the China Patent Database (CPD), encompassing some 2,837,000 numbered entries with application dates and descriptions. Next, stock trading data, corporate financials, and governance variables were sourced from the China Stock Market & Accounting Research (CSMAR) database. Finally, each firm's annual reports were collected from the Chinese Research Data Services (CNRDS) platform.

We further refined the sample firms according to the following screening criteria: (i) exclusion of firms under Special Treatment (ST) status to avoid the influence of financially distressed performers; (ii) removal of entities with significant missing or anomalous data; and (iii) winsorisation of all continuous variables at the 1st and 99th percentiles to avoid the impact of outliers. After integrating these databases and applying the above criteria, a final sample of 16,461 firm-year observations was obtained.

### 4.2 Measures

#### 4.2.1 Dependent variable: idiosyncratic risk (*IR*)

Following Aggarwal et al. (2011), Gu et al. (2019) and Aldawsari et al. (2025), we employ Fama and French's three-factor model (Fama & French, 1993) to calculate individual firms' idiosyncratic risk. First, we calculate the abnormal returns of firm *i* in month *m* by the following equation:

$$R_{i,m} = \alpha_i + \beta_{MKR} MKR_{i,m} + \beta_{HML} HML_{i,m} + \beta_{SMB} SMB_{i,m} + \varepsilon_{i,m} \quad (1)$$

wherein $R_{i,m}$ indicates stock returns of firm *i* that exceeds the risk-free rate on monthly basis. $MKR_{i,m}$ is the market risk premium factor that captures the monthly difference between Shanghai-Shenzhen Index and risk-free rate, while $HML_{i,m}$ is book-to-market risk premium factor that derives from differential returns between portfolios with high and low book-to-market ratios. $SMB_{i,m}$ represents the size-based factor, which is determined by the return differential between small and large-cap stock portfolios. While $\alpha_i$ represents the abnormal returns, $\beta_{MKR}$, $\beta_{HML}$ and $\beta_{SMB}$ demonstrate risk levels of each factor. Following Hudson and Morgan (2024) and Aldawsari et al. (2025), we leverage the residual term of $\varepsilon_{i,m}$ in Equation (1) to further compute firm i's IR in year t by executing the following equation:

$$IR_{i,t} = \sqrt{\frac{1}{12}\sum_{m=1}^{12}(\varepsilon_{i,m} - \varepsilon_{i,t})^2} \quad (2)$$

#### 4.2.2 Independent variable: Human-centric AI (*HCAI*)

Following Liu et al. (2025)' s work, we leverage pre-trained large language models (LLMs) to identify and quantify HCAI in this study. Specifically, we design a four-step process to execute the task of HCAI identification, which consists of learning context setting, prompt structuring, results evaluation and validation, and measuring construct.

***Step 1: Setting the learning context***. We leverage the grounding, bounding, and recasting constructs deriving from SAIT to build the HER framework of HCAI. This framework required sample firms' AI innovation to demonstrate the objectives of human-centred design, ethical management, and responsible implications. We then applied such the criteria and learning context to screen 2.873 million technological patents of sample firms.

***Step 2: Prompt Structuring***. To facilitate the identification process, we utilised a dynamic prompt engineering framework with specific inference parameters (temperature: 0.1, max_tokens: 150, top_p: 0.9) to prioritise deterministic and concise outputs (Doshi & Hauser, 2024). Structured prompts defined the LLM's role as a "Patent Innovation Identification Expert" incorporated classical AI definitions alongside our HCAI criteria, and enforced binary Yes/No outputs with brief explanations. Appendix A illustrate the specific structured prompt in this study.

***Step 3: Evaluating and validating results***.

We induce our test data and structured prompts through APIs among LLMs, including Google's Gemini 2.0 model, OpenAI's ChatGPT 4o model, Deepseek R1 model, and Alibaba's Qwen 2.5 model. As for the inference mode choice, we take Hillert et al. (2025)' suggestion to adopt few-shot approaches, which could enable LLMs to understand our context with few examples better. After running the structured prompts, we randomly extract 10,000 results and evaluate the prediction accuracy of each model. Table 1 presents the accuracy among four LLMs, implying that Qwen 2.5 achieved the highest performance in both training and test sets. Therefore, the optimised LLM performed automated identification across the patent corpus, with results validated through manual verification and expert evaluation, achieving high reliability confirmed by a Kappa score of 0.91. Figure 4 illustrates the detailed workflow of our HCAI identification process.

***Step 4: Measuring construct***. We operationalise $HCAI_{i,t}$ as the annual count of HCAI innovations from firm i in year t. To correct for substantial right-skewness in the distribution, we further took the logarithmic form of $\ln(1 + HCAI_{i,t})$ in our empirical tests.

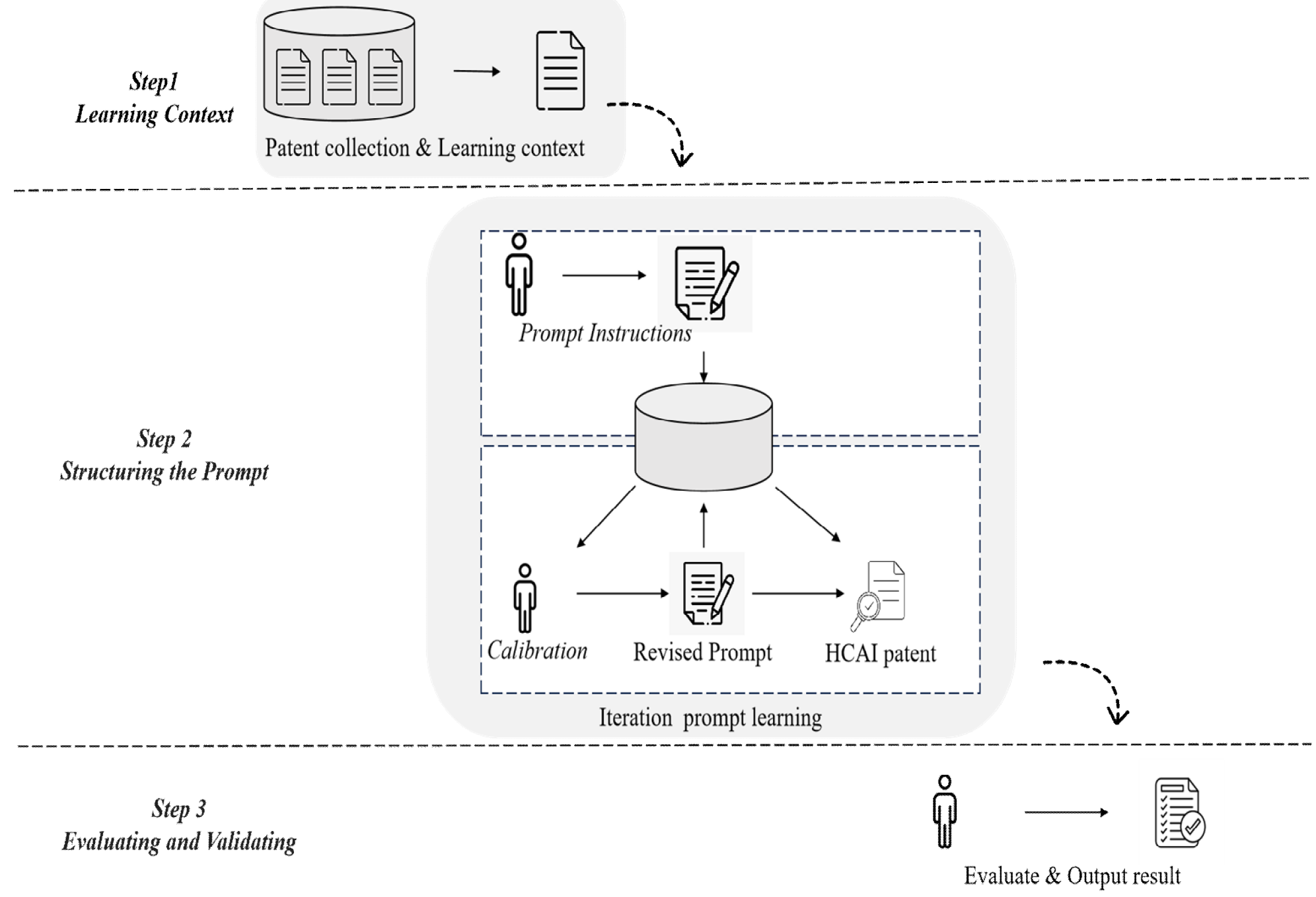

**Figure 4.** The procedures of applying LLM in HCAI identification

**[Insert Table 1 Here]**

While the LLM-based classification approach enables large-scale identification of HCAI-related innovations, it may inevitably involve a degree of classification noise due to ambiguity in patent language and overlapping technological concepts. However, such potential misclassification is unlikely to systematically bias our main results for several reasons. First, the identification framework is grounded in theoretically informed constructs derived from situated AI theory, ensuring conceptual consistency in screening criteria. Second, the high inter-model accuracy and the manually validated Kappa score of 0.91 indicate strong reliability of the classification outcomes. Third, any remaining noise is more likely to be random rather than directional, which would attenuate estimated coefficients toward zero rather than generate spurious significance. Therefore, the observed negative relationship between HCAI and idiosyncratic risk can be viewed as a conservative estimate of the true effect.

#### 4.2.3 Moderators

***Digitalisation (Digital)***: Following prior research on digitalisation (Qin et al., 2025; Tian et al., 2023), we quantify digital transformation using text analysis to extract relevant content from corporate annual reports. Figure 5 outlines the identification process. Specifically, we first obtained the annual report texts from the CNINFO for all sample firms during the observation period[5]. Using the Jieba tokeniser in Python, the texts are cleaned and structured. Next, we conduct text matching based on a seed dictionary of terms related to digital technologies, as these words represent the foundational elements of digital transformation strategies mentioned in annual reports. Following prior research, we constructed and expanded the relevant dictionary using keywords associated with "artificial intelligence," "big data," "cloud computing," and "internet technologies." Finally, we identified only those texts that describe the application scenarios of these digital technologies within business operations. This is because digital transformation emphasises the integration of digital technologies with business operations rather than merely pursuing technological digitalisation. For example, a representative disclosure from Midea Group's 2024 annual report states: "The company continuously deepens data-enabled business operations by leveraging 'Data + AI' technologies to deliver intelligent, data-driven solutions across core business scenarios including R&D, marketing, supply chain management, overseas markets, and ToB services" [6]. Such statements explicitly reflect the integration of digital technologies into core operational processes rather than merely referencing technological adoption. Through the above process, we aggregated the word frequencies from the digitalisation dictionary to obtain the total word frequency of digital transformation-related terms. This total frequency is then divided by the total number of words in the annual report and multiplied by 100 to calculate the final $Digital_{i,t}$ of firm $i$ in year $t$.

---

[5] Source: www.cninfo.com.cn

[6] Source available at: https://www.cninfo.com.cn/new/disclosure/detail?plate=szse&orgId=9900005965&stockCode=000333&announcementId=1222951181&announcementTime=2025-03-29

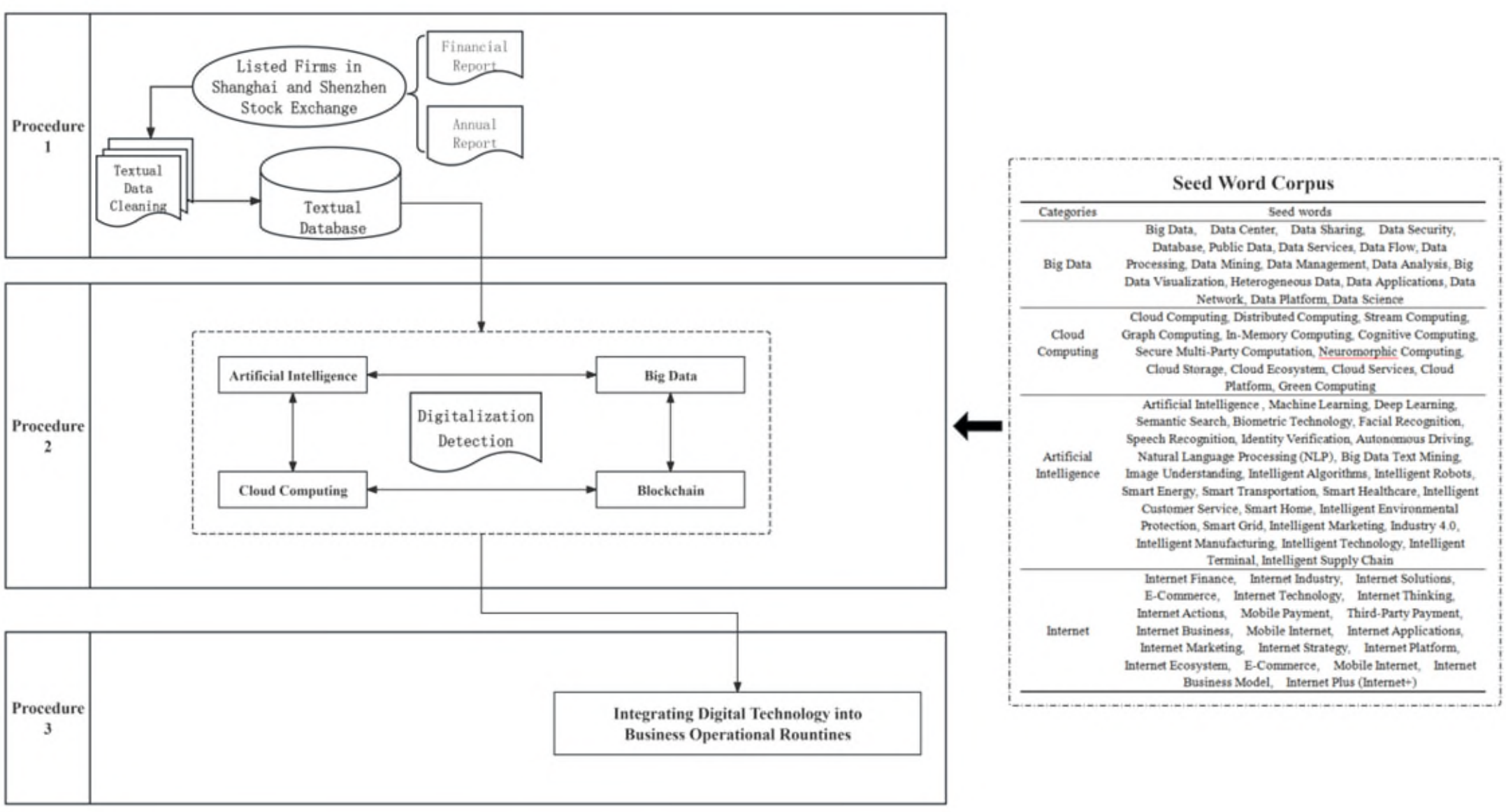

**Figure 5.** The quantifying procedure of digitalisation

***Operational efficiency (OE)***: Drawing on the approach of (Lam et al., 2016), we adopted stochastic frontier analysis (SFA) to estimate firms' operational efficiency. The first step involves specifying a conventional stochastic production function for firm *i* in industry *j* in year *t*, which incorporates both expected outputs and operational inputs as follows:

$$\begin{aligned} \ln\left(Operating\ Income_{ijt}\right) &= \beta_0 + \beta_1 \ln\ (Number\ of\ Employees_{ijt} + \beta_2 \ln\ (Cost\ of\ Goods\ Sold_{ijt}) \\ &+ \beta_3 \ln\ (Capital\ Expenditure_{ijt}) + \varepsilon_{ijt} - \gamma_{ijt} \end{aligned} \quad (3)$$

wherein $i$ and $t$ represent the firm and year, respectively. $Operating\ Income_{ijt}$ represents the outputs of firm i operating in industry j in year t, and the inputs for firms operations include $Number\ of\ Employees_{ijt}$, $Cost\ of\ Goods\ Sold_{ijt}$, and $Capital\ Expenditure_{ijt}$. While $\varepsilon_{ijt}$ delineates the random and unobserved factors that influence firms operations, $\gamma_{ijt}$ specifically refers to the inefficient items that ranges from 0 to 1. With value of 0, $\gamma_{ijt}$ indicates the firms achieve at the maximised frontier with no efficiency loss. Therefore, $\gamma_{ijt}$ is a relative measure of how inefficient a firm is when compared with the corresponding frontier in the same industry within the same year. In this sense, the operational efficiency of firm i in year t could be estimated as follows:

$$OE_{ijt} = 1 - \widehat{\gamma_{ijt}} \quad (4)$$

***Executive shareholding (Ehold)***: We follow Lyu et al. (2025) to identify directors, supervisors, board of supervisors, and senior managerial staff as the executives of sample firms. $Ehold_{it}$ is measured by the ratio of total shares held by these executives to the total number of shares issued by the sample firm i in year t. The higher the value of $Ehold_{it}$, the more substantial equity incentives the sample firms implement.

***IT background CEO (IT CEO)***: We first compiled sampled firms' CEO profiles from CNRDS, including both formal position holders and acting/intern CEO (e.g. deputy general managers). Second, we adopted the constructs from prior studies (Haislip & Richardson, 2018) to identify IT-related backgrounds from CEOs' educations, professional certifications and work experience. Finally, all screening results underwent manual verification to ensure accuracy, and thereby we constructed $IT\ CEO_{it}$.

#### 4.2.4 Control variables.

We control an extensive set of variables, including firm characteristics and corporate governance levels, that may affect sample firms' IR. Specifically, we first control firm size (*Size*), as large firms obtain more abundant resources to counter stock market risks (Hasan & Habib, 2017). Extensive prior evidences guide us to control firm age (*Age*), given that older companies deliver more information and media exposure that significantly reduces idiosyncratic volatility (Ferreira & Laux, 2007; Zhang, 2006). We follow Rajgopal and Venkatachalam (2011) to control firm leverage (*Lev*), which increases IR via enhancing the uncertainty of future cash flows. More recent studies on IR have further uncovered the effects of cash flow volatility (CashFlow), firm growth (Growth), firm financial performance (ROA), and the Book-to-market ratio (BM) on firms' IR by affecting cash flows (Gu et al., 2019). We hence also control all these variables in this study. As for corporate governance variances, we follow Hudson and Morgan (2024) and He et al. (2025) to control board size (*Bsize*), institutional ownership (*INST*), and top 1 ownership (*Top1*), constraining the governance structure and directors' influences on firms' IR. Table 2 illustrates all the variables' symbols, measurements and data sources in this study.

**[Insert Table 2 Here]**

### 4.3 Model specification

We employ a set of panel data regression models to testify proposed hypotheses. Specifically, we specify the effect of HCAI on firms' IR as well as the contingent factors in the HCAI-IR relationship through Equation (5) and Equation (6), respectively:

$$IR_{it} = \beta_0 + \beta_1 HCAI_{it} + \sum_{i=1}^{n} \beta_i Control_{it} + \sum Industry + \sum Year + \varepsilon_{it} \quad (5)$$

$$IR_{it} = \beta_0 + \beta_1 HCAI_{it} + \beta_2 Moderators_{it} + \beta_3 HCAI_{it} * Moderators_{it} + \sum_{i=1}^{n} \beta_i Control_{it} + \sum Industry + \sum Year + \varepsilon_{it} \quad (6)$$

wherein $IR_{it}$ represents the idiosyncratic risk of firm *i* in year *t*; $HCAI_{it}$ refers to the human-centric AI innovations of firm *i* in the observation year *t*; $Moderators_{it}$ is a vector containing all the contingent factors that may moderate the HCAI-IR relationship from both socio (i.e, $ITCEO_{it}$ and $TMS_{it}$) and technical (i.e.,$Digital_{it}$ and $OE_{it}$) subsystems. The vector of $Control_{it}$ includes all the variables for controlling firm characteristics and corporate governance that influence IR in this study. To address unobservable influences caused by time and industry, we also include the time- and industry-specific fixed effects via $\sum Year$ and $\sum Industry$, respectively. All regression analyses were performed in Stata 16.0 using the reghdfe function. Appendix B provide more details of stata coding we applied in this study.

## 5. Results

### 5.1 Descriptive statistics

Table 3 reports the descriptive statistics for all variables before regression analysis. To evaluate the potential issue of multicollinearity, we computed the variance inflation factors (VIFs) for all predictors in the model. The maximum VIF observed was 1.34, which is substantially below the conventional threshold of 10 , indicating no serious multicollinearity concerns among the independent variables (Aldawsari et al., 2025; Chu et al., 2024; Madan & Ashok, 2025).

**[Insert Table 3 Here]**

### 5.2 Hypothesis testing

Table 4 reports the regression results of the effect of HCAI on IR as well as the moderating effects of digitalisation, operational efficiency, executive shareholdings and IT background CEO, respectively. Specifically, the regression results in Column 1 of Table 4 indicate that HCAI has a significant negative effect on IR ($\beta = -0.024$, $p < 0.05$). Such a result implies that each logarithmic increase in HCAI innovation output reduces firms' IR by 2.4 percentage points, supporting H1 proposed in our study that HCAI significantly curbs firms' IR.

The analysis results in Columns 2 and 3 demonstrate how technical factors moderate the HCAI-IR relationship. Specifically, the coefficient of interaction item between HCAI and digitalisation ($\beta = -0.0766$, $p < 0.01$) shows that firms' digitalisation could strengthen the risk-mitigating effect of HCAI. Therefore, H2 was supported. However, the coefficient of interaction item between HCAI and operational efficiency ($\beta = 0.1814$, $p < 0.05$) in Column 3 suggests that HCAI efforts of firms with operational efficiency advantages significantly increase rather than reduce IR. Such a result profoundly rejects H3, implying that investors react negatively to HCAI innovations when firms are highly operationally efficient. This result diverges from prior findings that operational efficiency typically amplifies the benefits of technological innovation (Chuang et al., 2019; Li et al., 2021), pointing to the distinctive nature of HCAI.

Further, the analysis results in Columns 4 and 5 illustrate how factors embedded in firms' socio-subsystems moderate the effects of HCAI on IR. In particular, the interactive coefficient of HCAI and executive shareholding remains significantly negative ($\beta = 0.0001$, $p < 0.01$), suggesting that executive equity incentives could reinforce the risk-reducing effect of HCAI. On the other hand, IT background CEOs not only diminish but also convert the effect into risk-increasing ones (β = 0.0050, $p < 0.10$). This result indicates that investors react negatively to the HCAI innovation of firms with IT background CEOs. In this regard, our analysis results support H4 but reject H5.

**[Insert Table 4 Here]**

### 5.2 Robustness tests

We undertook alternative measurements, time-lagged regression, propensity score matching (PSM) regression, and sub-sample regression to check the robustness of the analysis results above. First, we follow He et al. (2022) and Hudson and Morgan (2024) to apply the Fama-French four-factor (FF-4) model to IR estimation. Compared with FF-3 model, FF-4 took the momentum factor $UMD_{im}$ into account when calculating stock returns of firm $i$ in month $m$, capturing the difference between high stock performance firms and low ones. The results in Column 1 of Table 5 represent the analysis results of an alternative measure of IR (β = -0.0026, $p < 0.05$), which remains consistent with our main regression model. Second, we also lagged IR by one observation period, given that HCAI may exist with a time-lag from innovation to implementation. The coefficient in Column 2 of Table 5 is β = -0.0036, $p < 0.01$, indicating that HCAI also reduced firms' IR lagged by one period, aligning with the main result.

**[Insert Table 5 Here]**

Third, we employed propensity score matching (PSM) to address potential endogeneity arising from sample selection bias (Ding et al., 2024). A dummy variable was constructed to classify firms into treatment and control groups. Specifically, firms with HCAI values above the sample mean of 0.031 were assigned to the treatment group (dummy variable value = 1). In contrast, those below the mean formed the control group (dummy variable value = 0). We then performed 1:1 nearest-neighbour matching without replacement, using the same control variables as in our main regression models. As reported in Table 6, balance tests confirm that all covariants in both treatment and control groups show no statistically significant differences in means after matching, indicating that the matching procedure achieved comparable groups.

Finally, we re-estimated our main regressions using the matched sample. The results are presented in Column 3 of Table 5, which is consistent with our baseline regression models and then reinforces the robustness of our findings.

**[Insert Table 6 Here]**

Finally, we followed Tian et al. (2023) to test the robustness of our moderating analysis results via sub-sample regression. Specifically, we divided our sample firms into four pair groups according to above or below the mean values of digitalisation (i.e., High-Digital group and Low-Digital group), operational efficiency (i.e., High-OE group and Low-OE group), executive shareholdings (i.e., High-Ehold group and Low-Ehold group), and IT background CEO (i.e., High-IT CEO group and Low-IT CEO group), respectively. Table 7 shows the sub-sample regression results, which remain the same as our main moderating analysis parts.

**[Insert Table 7 Here]**

Collectively, the analysis results above testified the robustness of main our finding through mitigating endogeneity concerns caused by sample selection bias, errors of variable measurement, reverse causality, and model sensitivity, confirming the negative relationship between HCAI and IR. Moreover, the moderation analysis results also maintain consistent with prior findings via sub-sample splits, reinforcing the robustness of boundary conditions among the HCAI-IR relationship.

## 6. Discussion and conclusion

This study aims to address how investors respond to firms' HCAI as well as what contingent factors influence the HCAI-IR relationship. Integrating situated AI theory with the socio-technical systems framework, we econometrically analysed a novel data set of A-share listed firms in China from 2015 to 2023. We found that HCAI reduces firms' idiosyncratic risk.

The cushioning role of HCAI is more substantial for firms with a higher level of digitalisation and executive shareholdings. However, the analysis results further reveal that HCAI experiences adverse investor reactions for firms with higher operational efficiency and more IT background CEOs, which is manifested in an increasing IR. Based on these findings, we generate several theoretical insights and practical implementations for both HCAI development and firm risk management during Industry 5.0 transition.

### 6.1 Theoretical contributions

First, we contribute to the ongoing HCAI discussions by revealing its curbing effects on firms' IR. Prior studies on HCAI lay their emphasis on either its conceptualisation and design or investigating how to integrate HCAI into intelligent manufacturing processes, all of which identify HCAI as the core driver for improving operational and social performance in Industry 5.0 transition. However, the existing literature overlooks how investors respond to HCAI of their portfolio firms. As one of the primary stakeholders, investors play a vital role in acquiring financial resources and managing firm risks. In this regard, this study explores the effect of HCAI on firms' IR, which has been extensively leveraged as an indicator of investor reactions across fields of information systems, accounting and finance, economics, and management studies. Even though prior studies also examine the same risk pattern at occupation, industry and firm exposure levels (Felten et al., 2021; Hudson & Morgan, 2024), we show that internal practices, design and governance sit closer to the risk outcome than exposure alone. Specifically, we find that HCAI situates firms' AI strategy to respond to the requirements of Industry 5.0, as it addresses AI instinct limitations through human-centred design, ethical management, and responsible implications. In this sense, HCAI aligns firms' AI desires with the broader interests of stakeholders, which lowers firms' IR by ensuring sustainable and stable profitability and cash flows during technological changes.

Secondly but more importantly, our findings also pinpoint that HCAI is not the panacea for lowering firm risks across all scenarios, given that firms' characteristics embedded in their socio and technical subsystems significantly moderate the HCAI-IR relationship. Although digitalisation and executive shareholding strengthen the risk-reducing effect of HCAI, we further found that operational efficiency and IT-background CEOs eliminate the risk-decreasing effect of HCAI. While extensive research demonstrates that operational efficiency amplifies the performance benefits of technological innovations (Lam et al., 2016; Li et al., 2021), we find that it unexpectedly weakens the risk-reducing effect of HCAI. This divergence suggests a boundary condition that efficiency magnifies gains when innovation is purely technical, but may backfire when innovation entails socio-technical reconfiguration. Also, HCAI needs cross-functional checks in technically led firms (Chuang et al., 2019; Li et al., 2021). In this regards, the contingent boundaries of the HCAI-IR relationship support external validity, suggesting that socio-technical alignment is more likely to support stability.

Third, we contribute to theoretical developments by integrating SAIT with STS theory to decode HCAI and derive financial outcomes that investors observe. Existing studies explain performance through the fit between AI, routines, structures, and people, with emphasis on efficiency and ethics (Kemp, 2024; Moser et al., 2024). We build on these studies and connect human-centric design and governance to a finance construct with established definitions, idiosyncratic risk estimated from asset-pricing residuals (Ang et al., 2006; Fama & French, 1993; Goyal & Santa-Clara, 2003; Xu & Malkiel, 2003). Moreover, we also distinguish HCAI from other established AI frameworks through the lens of SAIT. Although “trustworthy AI” or “responsible AI” has articulated the ethical principles AI systems should satisfy, these established concepts and frameworks remain largely principle-based and context-agnostic (Jobin et al., 2019). On the contrary, HCAI is firms’ internal situating approach, as HCAI embeds those principles into firms’ unique operational processes, governance mechanisms, and stakeholder relationship management, thereby answering how ethics can be operationalised through firms’ proactive design, ethical issue management, and deployment choices during the Industry 5.0 transition. Even though our contribution is incremental, we believe it has the theoretical potential to open a direct path for IS scholars to integrate AI design, AI management, and AI governance perspectives when further delving into financial risk management in the AI era.

Finally, we also provide an pragmatic measure of HCAI that is suitable for large-sample tests. The index is derived by classifying patent text with a large language model and a taxonomy of trustworthy and human-centric AI. It covers human-centred design, ethical management and responsible applications. We aggregate at the firm-year level (Harfouche et al., 2023; Heyder et al., 2023). The approach is transparent and replicable, and more targeted than broad AI intensity or exposure counts. It supports more direct tests of socio-technical claims, and it complements text-based approaches that infer AI engagement from public filings and patenting activity in innovation research (Ante & Saggu, 2025; Czarnitzki et al., 2023).

### 6.2 Managerial implications

This study offers distinct implications for corporate managers as well as investors and analysts. For corporate managers, our study indicates that HCAI could be updated into a firm's situating AI strategy that not only copes with technological turbulence but manages financial risks under the ongoing AI-driven transformation. To achieve these strategic goals, We further demonstrate the specific road map for managers to develop effective and idiosyncratic HCAI through three pathways, i.e., human-centric design, ethical management, and responsible implications.

Second but more importantly, we also inspire managers that the risk-curbing effect of HCAI is contingent on organizational socio-technical contexts. As the moderation evidence points to the stronger association in firms with higher digitalisation and executive shareholdings, managers could take more advantages of HCAI for risk management through both technical (i.e., improving digital foundations) and corporate governance (i.e., optimising shareholding incentives) approaches. This aligns with the special issue's focus on accountability and controls in organisational AI and with the socio-technical lens used in this study (Harfouche et al., 2023; Heyder et al., 2023; Kemp, 2024; Moser et al., 2024).

For investors and financial analysts, our results provide implications for evaluating corporate AI strategies beyond mere industrial exposure or adoption announcements. Investors and analysts should look for portfolio firms' HCAI clues through the evidences regarding human-centred design, ethical management, and responsible deployment practices, incorporating these contextual factors into their risk assessments and earnings forecasts. Moreover, they should pay more attention to firms with operational efficiency merits as well as IT background CEOs, as these factors embedded in firms' socio-technical systems hampers the bright side of HCAI.

### 6.3 Limitations and future directions

The main limitation of our study is that it relies on China A-share listed firms from 2015 to 2023. However, institutions, disclosure norms and investor behaviours differ across capital markets. External validity is bounded. As a result, replication in other contexts is needed to test the generalisability of the findings. The second limitation is measurement. Our outcome arises from the fact that the outcome focuses on idiosyncratic risk estimated from asset-pricing residuals. This captures firm volatility in returns, not operational incidents or systematic exposure. Monthly horizons and factor choices influence estimates.

Future research could focus on four directions. First, a follow-up study should adopt event-based and quasi-experimental designs, for example, policy shocks, governance announcements or incident disclosures. Second, future research could be conducted to triangulate the HCAI construct by combining patent text with annual reports, governance disclosures, model cards, audit trails, and incident logs, along with a manual coding audit and inter-rater checks. Third, future studies could broaden outcomes and settings. They could pair idiosyncratic risk with operational indicators such as incident rates, customer complaints, product recalls, and cybersecurity events. Also, they can test weekly versus monthly risk windows and alternative factor models, and extend the analysis to markets with different regulatory stringency and disclosure rules to probe scope conditions. Fourth, HCAI remains an evolving concept that has not yet attained universally accepted definitions. Future research could re-examine the HCAI-IR relationship by comparing our current and exploratory operationalisation with emerging connotations of HCAI as the construct matures over time.

## Declarations

**Ethics approval and consent to participate:** This research did not involve human participants and/or animals.

**Consent for publication:** The author(s) give their explicit consent that this research was not conducted in any organization/ company.

**Availability of data and material:** The datasets generated during and/or analyzed during this study are available from the corresponding author on reasonable request.

**Competing interests:** The author(s) declared no potential conflicts of interest with respect to the research, authorship, and/or publication of this article.

**Funding**: This research is funded by the Jiangsu Provincial Social Science Foundation (No. 25GLB022) and the Fundamental Research Funds for the Central Universities (NO. 2025SK09, China University of Mining and Technology).

**Authors' contributions:** All the authors have made equally substantial contributions to the conception, design of the work, the acquisition, analysis, and interpretation of data.

**Acknowledgement:** We are thankful to the reviewers and Editors for providing us with their valuable suggestions.

## References

Aggarwal, N., Dai, Q., & Walden, E. A. (2011). The more, the merrier? How the number of partners in a standard-setting initiative affects shareholder's risk and return1 [Article]. *MIS Quarterly: Management Information Systems*, *35*(2), 445-462. https://doi.org/10.2307/23044051

Aldawsari, H., Choudhry, T., & Luo, D. (2025). CEO power and firm risk at the onset of the 2007 financial crisis and the COVID-19 health crisis: international evidence. *Review of Quantitative Finance and Accounting*, *64*(4), 1633-1670. https://doi.org/10.1007/s11156-024-01347-4

Ancillai, C., Sabatini, A., Gatti, M., & Perna, A. (2023). Digital technology and business model innovation: A systematic literature review and future research agenda. *Technological Forecasting and Social Change*, *188*, 122307. https://doi.org/10.1016/j.techfore.2022.122307

Ang, A., Hodrick, R. J., Xing, Y., & Zhang, X. (2006). The cross-section of volatility and expected returns [Article]. *Journal of Finance*, *61*(1), 259-299. https://doi.org/10.1111/j.1540-6261.2006.00836.x

Ante, L., & Saggu, A. (2025). Quantifying a firm's AI engagement: Constructing objective, data-driven, AI stock indices using 10-K filings [Article]. *Technological Forecasting and Social Change*, *212*, Article 123965. https://doi.org/10.1016/j.techfore.2024.123965

Asness, C., Frazzini, A., Gormsen, N. J., & Pedersen, L. H. (2020). Betting against correlation: Testing theories of the low-risk effect [Article]. *Journal of Financial Economics*, *135*(3), 629-652. https://doi.org/10.1016/j.jfineco.2019.07.003

Bankins, S., Ocampo, A. C., Marrone, M., Restubog, S. L. D., & Woo, S. E. (2024). A multilevel review of artificial intelligence in organizations: Implications for organizational behavior research and practice [Review]. *Journal of Organizational Behavior*, *45*(2), 159-182. https://doi.org/10.1002/job.2735

Bansal, P., & Clelland, I. (2004). Talking trash: Legitimacy, impression management, and unsystematic risk in the context of the natural environment [Article]. *Academy of Management Journal*, *47*(1), 93-103. https://doi.org/10.2307/20159562

Bednar, P. M., & Welch, C. (2020). Socio-Technical Perspectives on Smart Working: Creating Meaningful and Sustainable Systems. *Information Systems Frontiers*, *22*(2), 281-298. https://doi.org/10.1007/s10796-019-09921-1

Bendig, D., Wagner, R., Piening, E. P., & Foege, J. N. (2023). ATTENTION TO DIGITAL INNOVATION: EXPLORING THE IMPACT OF A CHIEF INFORMATION OFFICER IN THE TOP MANAGEMENT TEAM [Article]. *MIS Quarterly: Management Information Systems*, *47*(7), 1487-1516. https://doi.org/10.25300/MISQ/2023/17152

Berente, N., Gu, B., Recker, J., & Santhanam, R. (2021). Managing Artificial Intelligence1. *Management Information Systems Quarterly*, *45*(3), 1433-1450. https://doi.org/10.25300/misq/2021/16274

Bernile, G., Bhagwat, V., & Yonker, S. (2018). Board diversity, firm risk, and corporate policies [Article]. *Journal of Financial Economics*, *127*(3), 588-612. https://doi.org/10.1016/j.jfineco.2017.12.009

Boido, C., & Aliano, M. (2025). Artificial intelligence exchange-traded funds: the intersection of finance, technology and sustainability [Article]. *Journal of Asset Management*, *26*(4), 345-354. https://doi.org/10.1057/s41260-025-00408-0

Bridoux, F., & Stoelhorst, J. W. (2022). STAKEHOLDER GOVERNANCE: SOLVING THE COLLECTIVE ACTION PROBLEMS IN JOINT VALUE CREATION [Article]. *Academy of Management Review*, *47*(2), 214-236. https://doi.org/10.5465/amr.2019.0441

Campbell, J. L., Chen, H., Dhaliwal, D. S., Lu, H. M., & Steele, L. B. (2014). The information content of mandatory risk factor disclosures in corporate filings [Article]. *Review of Accounting Studies*, *19*(1), 396-455. https://doi.org/10.1007/s11142-013-9258-3

Cao, Z., Feng, H., & Wiles, M. A. (2024). When Do Marketing Ideation Crowdsourcing Contests Create Shareholder Value? The Effect of Contest Design and Marketing Resource Factors. *Journal of Marketing*, *88*(2), 99-120. https://doi.org/doi: 10.1177/00222429231191446

Cath, C., Wachter, S., Mittelstadt, B., Taddeo, M., & Floridi, L. (2018). Artificial Intelligence and the 'Good Society': the US, EU, and UK approach [Article]. *Science and Engineering Ethics*, *24*(2), 505-528. https://doi.org/10.1007/s11948-017-9901-7

Chowdhury, S., Budhwar, P., & Wood, G. (2024). Generative Artificial Intelligence in Business: Towards a Strategic Human Resource Management Framework [Article]. *British Journal of Management*, *35*(4), 1680-1691. https://doi.org/10.1111/1467-8551.12824

Chu, X., Bai, Y., & Li, C. (2024). The Dark Side of Firms' Green Technology Innovation on Corporate Social Responsibility: Evidence from China. *Journal of Business Ethics*, *195*(1), 47-66. https://doi.org/10.1007/s10551-023-05556-0

Chuang, H. H. C., Oliva, R., & Heim, G. R. (2019). Examining the Link between Retailer Inventory Leanness and Operational Efficiency: Moderating Roles of Firm Size and Demand Uncertainty [Article]. *Production and Operations Management*, *28*(9), 2338-2364. https://doi.org/10.1111/poms.13055

Czarnitzki, D., Fernández, G. P., & Rammer, C. (2023). Artificial intelligence and firm-level productivity [Article]. *Journal of Economic Behavior and Organization*, *211*, 188-205. https://doi.org/10.1016/j.jebo.2023.05.008

Dai, J., Geng, R., Xu, D., Shangguan, W., & Shao, J. (2024). Unveiling the impact of the congruence between artificial intelligence and explorative learning on supply chain resilience. *International Journal of Operations & Production Management*, *45*(2), 570-593. https://doi.org/10.1108/ijopm-12-2023-0990

Dastin, J. (2022). Amazon Scraps Secret AI Recruiting Tool that Showed Bias against Women *. In *Ethics of Data and Analytics* (pp. 296-299). https://doi.org/10.1201/9781003278290-44

Deng, B., Peng, Z., Albitar, K., & Ji, L. (2025). Top management team stability

and ESG greenwashing: Evidence from China. *Business Strategy and the Environment*, *34*(1), 450-467. https://doi.org/10.1002/bse.3998

Ding, X. A., Sheng, Z. L., Appolloni, A., Shahzad, M., & Han, S. J. (2024). Digital transformation, ESG practice, and total factor productivity. *Business Strategy and the Environment*, *33*(5), 4547-4561. https://doi.org/10.1002/bse.3718

Do Khac, L. T., Leyer, M., Wichmann, J., & Richter, A. (2025). Divisional perspectives on trust in generative AI: an empirical study of senior decision-makers [Article]. *Enterprise Information Systems*, *19*(7-8), Article 2481557. https://doi.org/10.1080/17517575.2025.2481557

Doshi, A. R., & Hauser, O. P. (2024). Generative AI enhances individual creativity but reduces the collective diversity of novel content [Article]. *Science Advances*, *10*(28), Article adn5290. https://doi.org/10.1126/sciadv.adn5290

Eccles, R. G., Lee, L. E., & Stroehle, J. C. (2020). The Social Origins of ESG: An Analysis of Innovest and KLD [Article]. *Organization and Environment*, *33*(4), 575-596. https://doi.org/10.1177/1086026619888994

Eitel-Porter, R. (2021). Beyond the promise: implementing ethical AI [Article]. *AI and Ethics*, *1*(1), 73-80. https://www.scopus.com/inward/record.uri?eid=2-s2.0-85103234052&partnerID=40&md5=63341b1616b97f12aca8d4b48ee58f7e

Fama, E. F., & French, K. R. (1993). Common risk factors in the returns on stocks and bonds [Article]. *Journal of Financial Economics*, *33*(1), 3-56. https://doi.org/10.1016/0304-405X(93)90023-5

Felten, E., Raj, M., & Seamans, R. (2021). Occupational, industry, and geographic exposure to artificial intelligence: A novel dataset and its potential uses [Article]. *Strategic Management Journal*, *42*(12), 2195-2217. https://doi.org/10.1002/smj.3286

Ferreira, M. A., & Laux, P. A. (2007). Corporate Governance, Idiosyncratic Risk, and Information Flow. *The Journal of Finance*, *62*(2), 951-989. https://doi.org/https://doi.org/10.1111/j.1540-6261.2007.01228.x

Goyal, A., & Santa-Clara, P. (2003). Idiosyncratic Risk Matters! [Review]. *Journal of Finance*, *58*(3), 975-1008. https://doi.org/10.1111/1540-6261.00555

Gu, M., Jiang, G. J., & Xu, B. (2019). The role of analysts: An examination of the idiosyncratic volatility anomaly in the Chinese stock market. *Journal of Empirical Finance*, *52*, 237-254. https://doi.org/10.1016/j.jempfin.2019.03.007

Haislip, J. Z., & Richardson, V. J. (2018). The Effect of CEO IT Expertise on the Information Environment: Evidence from Earnings Forecasts and Announcements. *Journal of Information Systems*, *32*(2), 71-94. https://doi.org/10.2308/isys-51796

Hao, C. (2024). Artificial Intelligence and Operational Efficieny of Chinese Manufacturing Firms [Article]. *Int. Cognit. J.*, *1*, 28-39. https://www.scopus.com/inward/record.uri?eid=2-s2.0-105013491565&partnerID=40&md5=5ebf27b18bd23eb1c2fade68a02a47c4

Harfouche, A., Quinio, B., & Bugiotti, F. (2023). Human-Centric AI to Mitigate AI Biases: The Advent of Augmented Intelligence [Article]. *Journal of Global Information Management*, *31*(5). https://doi.org/10.4018/JGIM.331755

Hasan, M. M., & Habib, A. (2017). Firm life cycle and idiosyncratic volatility. *INTERNATIONAL REVIEW OF FINANCIAL ANALYSIS*, *50*, 164-175. https://doi.org/10.1016/j.irfa.2017.01.003

He, F., Du, H., Li, Y., & Hao, J. (2025). Make it Right: Regulatory Intervention in Managers' Misconduct and Corporate Risk [Article]. *Journal of Business Ethics*. https://doi.org/10.1007/s10551-025-05934-w

He, F., Qin, S., Liu, Y., & Wu, J. G. (2022). CSR and idiosyncratic risk: Evidence from ESG information disclosure [Article]. *Finance Research Letters*, *49*, Article 102936. https://doi.org/10.1016/j.frl.2022.102936

Herskovic, B., Kelly, B., Lustig, H., & Van Nieuwerburgh, S. (2016). The common factor in idiosyncratic volatility: Quantitative asset pricing implications [Article]. *Journal of Financial Economics*, *119*(2), 249-283. https://doi.org/10.1016/j.jfineco.2015.09.010

Heyder, T., Passlack, N., & Posegga, O. (2023). Ethical management of human-AI interaction: Theory development review [Review]. *Journal of Strategic Information Systems*, *32*(3), Article 101772. https://doi.org/10.1016/j.jsis.2023.101772

Hillert, A., Niessen-Ruenzi, A., & Ruenzi, S. (2025). Mutual Fund Shareholder Letters: Flows, Performance, and Managerial Behavior [Article]. *Management Science*, *71*(5), 4453-4473. https://doi.org/10.1287/mnsc.2021.03417

Hudson, K., & Morgan, R. E. (2024). Industry Exposure to Artificial Intelligence, Board Network Heterogeneity, and Firm Idiosyncratic Risk [Article]. *Journal of Management Studies*. https://doi.org/10.1111/joms.13127

Jarrahi, M. H., Kenyon, S., Brown, A., Donahue, C., & Wicher, C. (2023). Artificial intelligence: a strategy to harness its power through organizational learning [Article]. *Journal of Business Strategy*, *44*(3), 126-135. https://doi.org/10.1108/JBS-11-2021-0182

Jia, N., Luo, X., Fang, Z., & Liao, C. (2024). WHEN AND HOW ARTIFICIAL INTELLIGENCE AUGMENTS EMPLOYEE CREATIVITY [Article]. *Academy of Management Journal*, *67*(1), 5-32. https://doi.org/10.5465/amj.2022.0426

Jobin, A., Ienca, M., & Vayena, E. (2019). Artificial Intelligence: the global landscape of ethics guidelines. In *Nat. Mach. Intell. (2019)*.

Kemp, A. (2024). COMPETITIVE ADVANTAGE THROUGH ARTIFICIAL INTELLIGENCE: TOWARD A THEORY OF SITUATED AI [Article]. *Academy of Management Review*, *49*(3), 618-635. https://doi.org/10.5465/amr.2020.0205

Koeppen, M., Chen, M. Y., & Wu, H. T. (2025). Omnipresent AI and intelligent hyperautomation: methodologies, interactivity and applications [Review]. *Enterprise Information Systems*, Article 2551727. https://doi.org/10.1080/17517575.2025.2551727

Lam, H. K. S., Yeung, A. C. L., & Cheng, T. C. E. (2016). The impact of firms' social media initiatives on operational efficiency and innovativeness [Article]. *Journal of Operations Management*, *47-48*, 28-43. https://doi.org/10.1016/j.jom.2016.06.001

Lee, Y. S., Kim, T., Choi, S., & Kim, W. (2022). When does AI pay off? AI -

adoption intensity, complementary investments, and R&D strategy [Article]. *Technovation*, *118*, Article 102590. https://doi.org/10.1016/j.technovation.2022.102590

Li, B., Rajgopal, S., & Venkatachalam, M. (2014). R2 and idiosyncratic risk are not interchangeable [Article]. *ACCOUNTING REVIEW*, *89*(6), 2261-2295. https://doi.org/10.2308/accr-50826

Li, G., Li, N., & Sethi, S. P. (2021). Does CSR Reduce Idiosyncratic Risk? Roles of Operational Efficiency and AI Innovation [Article]. *Production and Operations Management*, *30*(7), 2027-2045. https://doi.org/10.1111/poms.13483

Li, J., Ma, S., Qu, Y., & Wang, J. (2023). The impact of artificial intelligence on firms' energy and resource efficiency: Empirical evidence from China [Article]. *Resources Policy*, *82*, Article 103507. https://doi.org/10.1016/j.resourpol.2023.103507

Lin, Y. M., & Shen, C. A. (2015). Family firms' credit rating, idiosyncratic risk, and earnings management [Article]. *Journal of Business Research*, *68*(4), 872-877. https://doi.org/10.1016/j.jbusres.2014.11.044

Liu, Z. Y. R., Dong, S. K., Zeng, W., Wang, Y. T., & Niu, D. F. (2025). Exploring the impact of human-centred AI on firms' social and operational performance: A large language model approach [Article]. *Transportation Research Part E: Logistics and Transportation Review*, *203*, Article 104381. https://doi.org/10.1016/j.tre.2025.104381

Lyu, B., Wang, H., Zhou, Y., & Wang, C. (2025). Does executive equity incentive boost up enterprise digital transformation? -The moderating role of strategic decision-making situation. *Asia Pacific Journal of Management*, 1-36. https://doi.org/10.1007/s10490-025-10039-z

Madan, R., & Ashok, M. (2025). Making Sense of AI Benefits: A Mixed-method Study in Canadian Public Administration. *Information Systems Frontiers*, *27*(3), 889-923. https://doi.org/10.1007/s10796-024-10475-0

Marcus, G. (2018). Deep Learning: A Critical Appraisal. *undefined*. https://doi.org/10.48550/ARXIV.1801.00631

Marioni, L. D. S., Rincon-Aznar, A., & Venturini, F. (2024). Productivity performance, distance to frontier and AI innovation: Firm-level evidence from Europe [Article]. *Journal of Economic Behavior and Organization*, *228*, Article 106762. https://doi.org/10.1016/j.jebo.2024.106762

Mathew, D., Brintha, N. C., & Jappes, J. T. W. (2023). Artificial Intelligence Powered Automation for Industry 4.0. In A. Nayyar, M. Naved, & R. Rameshwar (Eds.), *New Horizons for Industry 4.0 in Modern Business* (pp. 1-28). Springer International Publishing. https://doi.org/10.1007/978-3-031-20443-2_1

Matthews, M. J., Su, R., Yonish, L., McClean, S., Koopman, J., & Yam, K. C. (2025). A Review of Artificial Intelligence, Algorithms, and Robots Through the Lens of Stakeholder Theory [Review]. *Journal of Management*, *51*(6), 2627-2676. https://doi.org/10.1177/01492063241311855

Miric, M., Jia, N., & Huang, K. G. (2023). Using supervised machine learning for large-scale classification in management research: The case for identifying artificial intelligence patents. *Strategic Management Journal*, *44*(2), 491-519. https://doi.org/10.1002/smj.3441

Moon, S., Tuli, K. R., & Mukherjee, A. (2023). Does Disclosure of Advertising Spending Help Investors and Analysts? *Journal of Marketing*, *87*(3), 359-382. https://doi.org/10.1177/00222429221123013

Morgeson, F. V., Sharma, U., Schultz, X. W., Pansari, A., Ruvio, A., & Hult, G. T. M. (2024). Weathering the crash: Do customer-company relationships pay off during economic crises? *Journal of the Academy of Marketing Science*, *52*(2), 489-511. https://doi.org/10.1007/s11747-023-00947-1

Moser, C., Glaser, V. L., & Lindebaum, D. (2024). TAKING SITUATEDNESS SERIOUSLY IN THEORIZING ABOUT COMPETITIVE ADVANTAGE THROUGH ARTIFICIAL INTELLIGENCE: A RESPONSE TO KEMP'S "COMPETITIVE ADVANTAGES THROUGH ARTIFICIAL INTELLIGENCE" [Article]. *Academy of Management Review*, *49*(3), 683-685. https://doi.org/10.5465/amr.2023.0265

Nahavandi, S. (2019). Industry 5.0-a human-centric solution [Article]. *Sustainability (Switzerland)*, *11*(16), Article 4371. https://doi.org/10.3390/su11164371

Nguyen, H., Calantone, R., & Krishnan, R. (2020). Influence of social media emotional word of mouth on institutional investors' decisions and firm value [Article]. *Management Science*, *66*(2), 887-910. https://doi.org/10.1287/mnsc.2018.3226

Passalacqua, M., Pellerin, R., Doyon-Poulin, P., Del-Aguila, L., Boasen, J., & Léger, P. M. (2022). Human-Centred AI in the Age of Industry 5.0: A Systematic Review Protocol. In J. Y. Chen, G. Fragomeni, H. Degen, & S. Ntoa (Eds.), (Vol. 13518 LNCS, pp. 483-492): Springer Science and Business Media Deutschland GmbH.

Passalacqua, M., Pellerin, R., Magnani, F., Doyon-Poulin, P., Del-Aguila, L., Boasen, J., & Léger, P. M. (2025). Human-centred AI in industry 5.0: a systematic review [Review]. *International Journal of Production Research*, *63*(7), 2638-2669. https://doi.org/10.1080/00207543.2024.2406021

Psarommatis, F., May, G., & Azamfirei, V. (2023). Envisioning maintenance 5.0: Insights from a systematic literature review of Industry 4.0 and a proposed framework [Review]. *Journal of Manufacturing Systems*, *68*, 376-399. https://doi.org/10.1016/j.jmsy.2023.04.009

Qin, L.-b., Zeng, W., Liu, Z.-y., Dong, S., & Tse, Y. K. (2025). Navigating across the uncertainty: investigating the impact of buyer firms' digital transformation on operational efficiency. *INDUSTRIAL MANAGEMENT & DATA SYSTEMS*, 1-25. https://doi.org/10.1108/imds-03-2025-0292

Rajgopal, S., & Venkatachalam, M. (2011). Financial reporting quality and idiosyncratic return volatility. *Journal of Accounting and Economics*, *51*(1), 1-20. https://doi.org/10.1016/j.jacceco.2010.06.001

Reber, B., Gold, A., & Gold, S. (2022). ESG Disclosure and Idiosyncratic Risk in Initial Public Offerings [Article]. *Journal of Business Ethics*, *179*(3), 867-886. https://doi.org/10.1007/s10551-021-04847-8

Srinivasan, S., & Hanssens, D. M. (2009). Marketing and firm value: Metrics,

methods, findings, and future directions [Article]. *Journal of Marketing Research*, *46*(3), 293-312. https://doi.org/10.1509/jmkr.46.3.293

Sun, X., & Song, Y. (2025). Unlocking the Synergy: Increasing productivity through Human-AI collaboration in the industry 5.0 Era [Article]. *Computers and Industrial Engineering*, *200*, Article 110657. https://doi.org/10.1016/j.cie.2024.110657

Tan, M., & Liu, B. (2016). CEO's managerial power, board committee memberships and idiosyncratic volatility [Article]. *INTERNATIONAL REVIEW OF FINANCIAL ANALYSIS*, *48*, 21-30. https://doi.org/10.1016/j.irfa.2016.09.003

Tantalo, C., & Priem, R. L. (2016). Value creation through stakeholder synergy [Article]. *Strategic Management Journal*, *37*(2), 314-329. https://doi.org/10.1002/smj.2337

Tian, M., Chen, Y., Tian, G., Huang, W., & Hu, C. (2023). The role of digital transformation practices in the operations improvement in manufacturing firms: A practice-based view. In M. Tian, Y. Chen, G. Tian, W. Huang, & C. Hu (Eds.), *International Journal of Production Economics* (Vol. 262).

Tong, X., Linderman, K., & Zhu, Q. (2023). Managing a portfolio of environmental projects: Focus, balance, and environmental management capabilities. *Journal of Operations Management*, *69*(1), 127-158. https://doi.org/10.1002/joom.1201

Vesnic-Alujevic, L., Nascimento, S., & Pólvora, A. (2020). Societal and ethical impacts of artificial intelligence: Critical notes on European policy frameworks [Article]. *Telecommunications Policy*, *44*(6), Article 101961. https://doi.org/10.1016/j.telpol.2020.101961

Wamba-Taguimdje, S. L., Fosso Wamba, S., Kala Kamdjoug, J. R., & Tchatchouang Wanko, C. E. (2020). Influence of artificial intelligence (AI) on firm performance: the business value of AI-based transformation projects [Article]. *Business Process Management Journal*, *26*(7), 1893-1924. https://doi.org/10.1108/BPMJ-10-2019-0411

Wang, J., Liu, Y., Wang, W., & Wu, H. (2023). The effects of "machine replacing human" on carbon emissions in the context of population aging – Evidence from China [Article]. *Urban Climate*, *49*, Article 101519. https://doi.org/10.1016/j.uclim.2023.101519

Wang, K., Zhang, P., He, J., & Feng, C. (2024). Does corporate social responsibility reduce idiosyncratic risk? The contingent role of opportunity costs. *Asia Pacific Journal of Management*. https://doi.org/10.1007/s10490-024-10005-1

Wang, Q., Lau, R. Y. K., & Yang, K. (2020). Does the interplay between the personality traits of CEOs and CFOs influence corporate mergers and acquisitions intensity? An econometric analysis with machine learning-based constructs. *Decision Support Systems*, *139*, 113424. https://doi.org/10.1016/j.dss.2020.113424

Xu, K., Geng, C., Wei, X., & Jiang, H. (2020). Financing development, financing constraint and R&D investment of strategic emerging industries in China [Article]. *Journal of Business Economics and Management*, *21*(4), 1010-1034.

https://doi.org/10.3846/jbem.2020.12727

Xu, W., Dainoff, M. J., Ge, L., & Gao, Z. (2023). Transitioning to Human Interaction with AI Systems: New Challenges and Opportunities for HCI Professionals to Enable Human-Centered AI [Article]. *International Journal of Human-Computer Interaction*, *39*(3), 494-518. https://doi.org/10.1080/10447318.2022.2041900

Xu, X., Lu, Y., Vogel-Heuser, B., & Wang, L. (2021). Industry 4.0 and Industry 5.0—Inception, conception and perception [Article]. *Journal of Manufacturing Systems*, *61*, 530-535. https://doi.org/10.1016/j.jmsy.2021.10.006

Xu, Y., & Malkiel, B. G. (2003). Investigating the Behavior of Idiosyncratic Volatility [Article]. *Journal of Business*, *76*(4), 613-644. https://doi.org/10.1086/377033

Yao, N., Bai, J., Yu, Z., & Guo, Q. (2025). Does AI orientation facilitate operational efficiency? A contingent strategic orientation perspective [Article]. *Journal of Business Research*, *186*, Article 114994. https://doi.org/10.1016/j.jbusres.2024.114994

Zhang, X., Zhang, W., Wang, J., & Chen, B. (2025). Impact of CEO's IT Background on Green Technology Innovation: Evidence From China. *Corporate Governance: An International Review*, *33*(5), 1203-1222. https://doi.org/10.1111/corg.12644

Zhang, X. F. (2006). Information Uncertainty and Stock Returns. *The Journal of Finance*, *61*(1), 105-137. https://doi.org/10.1111/j.1540-6261.2006.00831.x

**Table 1.** Comparative performance of LLMs for HCAI identification

| Models | Training Set | Test Set | | |
|---|---|---|---|---|
| | Accuracy | Accuracy | Accuracy (Macro) | Accuracy (Weighted) |
| **Baseline** | | | | |
| Manual Annotation | 0.910 | 0.821 | 0.863 | 0.875 |
| **LLMs** | | | | |
| Gemini 2.0 | 0.746 | 0.755 | 0.749 | 0.748 |
| ChatGPT 4o | 0.896 | 0.898 | 0.878 | 0.891 |
| DeepSeek R1 | 0.912 | 0.893 | 0.903 | 0.916 |
| Qwen 2.5 | **0.928** | **0.953** | **0.916** | **0.927** |

**Table 2.** Variable definitions and measures

| | Symbol | Measurement | Data Sources |
|---|---|---|---|
| ***Dependent Variables*** | | | |
| Idiosyncratic Risk | IR | Fama-French three-factor (FF-3) | CSMAR |
| ***Independent variable*** | | | |
| Human-Centric AI | HCAI | Ln(HCAI_Number+1) | CPD |
| ***Moderators*** | | | |
| Digitalisation | Digital | Digital Transformation Index | CNRDS |
| Operational Efficiency | OE | Utilizing Stochastic frontier analysis to calculate | CSMAR |
| Executive Shareholding | Ehold | DSSE Shareholding Ratio | CSMAR |
| IT background CEO | IT CEO | Proportion of Technically-Backgrounded Executives | CSMAR |
| ***Control Variables*** | | | |
| Firm size | Size | Ln(firms' total assets) | CSMAR |
| Return on Assets | Roa | Return on Assets after tax | CSMAR |
| Financial Leverage | Lev | Debt to asset ratio | CSMAR |
| Firm age | Age | Number of years since a firm was publicly listed | CSMAR |
| Capital Intensity | Capital | Fixed Assets per Employee Ratio | CSMAR |
| Cash Flow Volatility | CashFlow | Three-Year Volatility of Cash Flow to Total Assets Ratio | CSMAR |
| Book-to-Market Ratio | BM | Company's Book Value to Market Value Ratio (×100%) | CSMAR |
| Firm growth | Growth | Revenue Growth Rate | CSMAR |
| Board size | Bsize | Natural Logarithm of the Number of Directors | CSMAR |
| The top1 ownership | Top1 | Largest Shareholder's Ownership Percentage | CSMAR |
| Institutional Ownership | INST | Institutional Investor Shareholding Ratio | CSMAR |

**Table 3.** Descriptive statistics and correlations

| | IR | HCAI | Digital | OE | Ehold | IT CEO | Size | Roa | Lev | Age | Capital | CashFlow | BM | Growth | Bsize | Top1 | INST |
|---|---|---|---|---|---|---|---|---|---|---|---|---|---|---|---|---|---|
| N | 16461 | 16461 | 16461 | 16461 | 16461 | 16461 | 16461 | 16461 | 16461 | 16461 | 16461 | 16461 | 16461 | 16461 | 16461 | 16461 | 16461 |
| mean | 0.085 | 0.031 | 0.02 | 0.613 | 16.52 | 0.73 | 22.34 | 0.036 | 0.406 | 2.153 | 2.281 | 0.04 | 0.343 | 0.147 | 2.1 | 0.325 | 0.333 |
| sd | 0.027 | 0.222 | 0.031 | 0.0631 | 19.69 | 0.444 | 1.22 | 0.073 | 0.185 | 0.682 | 2.749 | 0.057 | 0.155 | 0.334 | 0.194 | 0.142 | 0.275 |
| min | 0.034 | 0 | 0 | 0 | 0 | 0 | 20.07 | -0.965 | 0.057 | 0.693 | 0.095 | 0 | 0.005 | -0.569 | 1.099 | 0.021 | 0 |
| p50 | 0.082 | 0 | 0.01 | 0.034 | 6.159 | 1 | 22.14 | 0.037 | 0.403 | 2.197 | 1.89 | 0.031 | 0.324 | 0.099 | 2.197 | 0.301 | 0.315 |
| max | 0.185 | 4.663 | 0.363 | 1 | 127.4 | 1 | 26.36 | 0.759 | 0.887 | 3.526 | 212.7 | 5.915 | 1.207 | 2.086 | 2.833 | 0.9 | 0.91 |
| IR | 1 | | | | | | | | | | | | | | | | |
| HCAI | -0.019** | 1 | | | | | | | | | | | | | | | |
| Digital | 0.086*** | 0.131*** | 1 | | | | | | | | | | | | | | |
| OE | -0.005 | -0.002 | 0.008 | 1 | | | | | | | | | | | | | |
| Ehold | 0.137*** | -0.047*** | 0.065*** | -0.006 | 1 | | | | | | | | | | | | |
| IT CEO | 0.143*** | 0.079*** | 0.403*** | -0.001 | 0.123*** | 1 | | | | | | | | | | | |
| Size | -0.224*** | 0.193*** | -0.025*** | 0.008 | -0.361*** | -0.143*** | 1 | | | | | | | | | | |
| Roa | 0.011 | 0.011 | -0.057*** | -0.001 | 0.088*** | -0.029*** | 0.064*** | 1 | | | | | | | | | |
| Lev | -0.021*** | 0.067*** | -0.025*** | -0.002 | -0.249*** | -0.085*** | 0.485*** | -0.302*** | 1 | | | | | | | | |
| Age | -0.195*** | 0.078*** | -0.048*** | 0.011 | -0.515*** | -0.146*** | 0.489*** | -0.095*** | 0.284*** | 1 | | | | | | | |
| Capital | 0.012 | -0.020** | 0.01 | 0.01 | 0.027*** | 0.022*** | -0.032*** | -0.149*** | -0.084*** | -0.062*** | 1 | | | | | | |
| CashFlow | 0.065*** | -0.008 | -0.026*** | 0.002 | 0.026*** | -0.015* | -0.052*** | 0.019** | 0.007 | -0.038*** | 0.056*** | 1 | | | | | |
| BM | -0.397*** | -0.030*** | -0.047*** | 0.020*** | 0.067*** | -0.102*** | 0.045*** | 0.019** | -0.406*** | -0.001 | 0.062*** | -0.034*** | 1 | | | | |
| Growth | 0.172*** | -0.009 | -0.01 | -0.001 | 0.054*** | 0.014* | 0.049*** | 0.281*** | 0.047*** | -0.097*** | -0.026*** | 0.037*** | -0.130*** | 1 | | | |
| Bsize | -0.108*** | 0.047*** | -0.056*** | -0.005 | -0.221*** | -0.123*** | 0.261*** | 0.035*** | 0.125*** | 0.215*** | -0.017** | -0.048*** | 0.014* | 0.005 | 1 | | |
| Top1 | -0.096*** | -0.017** | -0.135*** | -0.002 | -0.069*** | -0.116*** | 0.168*** | 0.141*** | 0.014* | -0.057*** | -0.035*** | -0.003 | 0.063*** | 0.003 | -0.007 | 1 | |
| INST | 0.016** | 0.042*** | -0.103*** | -0.011 | -0.511*** | -0.127*** | 0.339*** | 0.131*** | 0.137*** | 0.182*** | -0.058*** | -0.023*** | -0.103*** | 0.098*** | 0.206*** | 0.375*** | 1 |

**Table 4.** Regression results

| | **H(1)** | **H(2)** | **H(3)** | **H(4)** | **H(5)** |
|---|---|---|---|---|---|
| | IR | IR | IR | IR | IR |
| HCAI | -0.0024** | -0.0007 | 0.0112** | 0.0011 | -0.0067** |
| | (-2.0164) | (-0.5046) | (2.4577) | (0.6720) | (-2.4434) |
| Digital | | 0.0145* | | | |
| | | (1.8088) | | | |
| ***HCAI*Digital*** | | **-0.0766*****| | | |
| | | **(-2.9547)** | | | |
| OE | | | -0.0004 | | |
| | | | (-0.3552) | | |
| ***HCAI_OE*** | | | **0.1814**** | | |
| | | | **(2.5499)** | | |
| Ehold | | | | 0.0001*** | |
| | | | | (6.9163) | |
| ***HCAI*Ehold*** | | | | **-0.0002***** | |
| | | | | **(-2.8048)** | |
| IT CEO | | | | | 0.0011*** |
| | | | | | (2.9233) |
| ***HCAI*IT CEO*** | | | | | **0.0050*** |
| | | | | | **(1.6680)** |
| Size | -0.0013*** | -0.0015*** | -0.0013*** | -0.0014*** | -0.0013*** |
| | (-6.1049) | (-6.9010) | (-6.1833) | (-6.3021) | (-6.0684) |
| Roa | -0.0317*** | -0.0336*** | -0.0316*** | -0.0315*** | -0.0315*** |
| | (-8.3064) | (-8.7849) | (-8.2924) | (-8.2486) | (-8.2645) |
| Lev | -0.0212*** | -0.0204*** | -0.0212*** | -0.0212*** | -0.0211*** |
| | (-13.7132) | (-13.1238) | (-13.6995) | (-13.7052) | (-13.6504) |
| Age | -0.0045*** | -0.0035*** | -0.0045*** | -0.0045*** | -0.0044*** |
| | (-15.2502) | (-10.6802) | (-15.2642) | (-15.1679) | (-15.1033) |
| Capital | 0.0008*** | 0.0008*** | 0.0008*** | 0.0008*** | 0.0008*** |
| | (4.9958) | (5.1451) | (5.0841) | (5.0567) | (5.0150) |
| CashFlow | 0.0422*** | 0.0423*** | 0.0421*** | 0.0417*** | 0.0426*** |
| | (7.0951) | (7.1200) | (7.0802) | (7.0145) | (7.1662) |
| BM | -0.0725*** | -0.0720*** | -0.0726*** | -0.0726*** | -0.0724*** |
| | (-49.6641) | (-49.2206) | (-49.6825) | (-49.7306) | (-49.5675) |
| Growth | 0.0131*** | 0.0130*** | 0.0131*** | 0.0131*** | 0.0131*** |
| | (22.3503) | (22.1478) | (22.3608) | (22.3394) | (22.3221) |
| Bsize | -0.0035*** | -0.0033*** | -0.0035*** | -0.0035*** | -0.0037*** |
| | (-3.7104) | (-3.5847) | (-3.7295) | (-3.7109) | (-3.9456) |
| Top1 | -0.0069*** | -0.0088*** | -0.0069*** | -0.0069*** | -0.0068*** |
| | (-5.0999) | (-6.3426) | (-5.0783) | (-5.0396) | (-5.0363) |
| INST | -0.0007 | 0.0040*** | -0.0007 | -0.0007 | -0.0007 |
|  | (-0.8096) | (3.4819) | (-0.8011) |  |  |
| _cons | 0.1770*** | 0.1616*** | 0.1632*** | 0.1635*** | 0.1621*** |
|  | (36.3776) | (37.1110) | (37.4684) | (37.5220) | (37.1855) |
| *N* | 16461 | 16461 | 16461 | 16461 | 16461 |
| r2_a | 0.4165 | 0.4184 | 0.4166 | 0.4168 | 0.4168 |

*t* statistics in parentheses. * $p < 0.1$, ** $p < 0.05$, *** $p < 0.01$

**Table 5.** Robustness test results

| | Alternative Measure | Lagged by One Period | PSM |
|---|---|---|---|
| | IR_4 | IR | IR |
| HCAI | -0.0026** | -0.0036*** | -0.0066** |
| | (-2.2460) | (-3.2035) | (-2.2546) |
| Size | -0.0013*** | -0.0012*** | -0.0054*** |
| | (-6.1044) | (-5.3253) | (-4.2710) |
| Roa | -0.0317*** | -0.0322*** | -0.0048 |
| | (-8.2647) | (-8.1197) | (-0.1522) |
| Lev | -0.0206*** | -0.0211*** | 0.0015 |
| | (-13.4278) | (-13.1491) | (0.1087) |
| Age | -0.0044*** | -0.0046*** | -0.0026 |
| | (-15.0697) | (-15.1415) | (-1.0170) |
| Capital | 0.0008*** | 0.0008*** | -0.0010 |
| | (5.0290) | (4.7117) | (-0.6682) |
| CashFlow | 0.0418*** | 0.0443*** | 0.1369** |
| | (7.0458) | (7.1542) | (2.3358) |
| BM | -0.0706*** | -0.0723*** | -0.0486*** |
| | (-48.3946) | (-47.9509) | (-4.6360) |
| Growth | 0.0130*** | 0.0134*** | 0.0182*** |
| | (22.0179) | (21.6267) | (3.4118) |
| Bsize | -0.0031*** | -0.0036*** | 0.0030 |
| | (-3.4237) | (-3.6774) | (0.4966) |
| Top1 | -0.0068*** | -0.0072*** | -0.0220* |
| | (-5.0864) | (-5.1588) | (-1.8680) |
| INST | -0.0009 | -0.0004 | 0.0182** |
| | (-1.0162) | (-0.4923) | (2.2658) |
| _cons | 0.1706*** | 0.1727*** | 0.2231*** |
| | (35.5470) | (33.4157) | (8.5517) |
| *N* | 16461 | 15342 | 12370 |
| r2_a | 0.4117 | 0.4182 | 0.4404 |

*t* statistics in parentheses, * $p < 0.1$, ** $p < 0.05$, *** $p < 0.01$

**Table 6.** Balance test and propensity score matching results

| Variables | Unmatched | Mean | | %Bias | t-test | |
|---|---|---|---|---|---|---|
| | Matched | Treated | Control | | t | p>\|t\| |
| Size | U | 24.270 | 22.321 | 137.000 | 22.080 | 0.000 |
| | M | 24.270 | 24.248 | 1.500 | 0.130 | 0.896 |
| Roa | U | 0.042 | 0.037 | 7.900 | 0.990 | 0.321 |
| | M | 0.042 | 0.041 | 0.600 | 0.050 | 0.959 |
| Lev | U | 0.513 | 0.405 | 62.700 | 8.010 | 0.000 |
| | M | 0.513 | 0.505 | 4.600 | 0.420 | 0.674 |
| Age | U | 2.638 | 2.147 | 74.400 | 9.840 | 0.000 |
| | M | 2.638 | 2.591 | 7.100 | 0.740 | 0.462 |
| Capital | U | 1.733 | 2.231 | -40.500 | -5.130 | 0.000 |
| | M | 1.733 | 1.747 | -1.100 | -0.130 | 0.896 |
| CashFlow | U | 0.035 | 0.039 | -15.700 | -2.000 | 0.045 |
| | M | 0.035 | 0.035 | -1.000 | -0.090 | 0.925 |
| BM | U | 0.302 | 0.343 | -28.100 | -3.700 | 0.000 |
| | M | 0.302 | 0.304 | -1.800 | -0.180 | 0.856 |
| Growth | U | 0.119 | 0.145 | -9.200 | -1.100 | 0.270 |
| | M | 0.119 | 0.140 | -7.300 | -0.790 | 0.427 |
| Bsize | U | 2.173 | 2.099 | 35.900 | 5.250 | 0.000 |
| | M | 2.173 | 2.168 | 2.500 | 0.240 | 0.811 |
| Top1 | U | 0.311 | 0.325 | -9.700 | -1.390 | 0.165 |
| | M | 0.311 | 0.309 | 0.900 | 0.090 | 0.930 |
| INST | U | 0.453 | 0.332 | 42.700 | 6.000 | 0.000 |
| | M | 0.453 | 0.436 | 5.700 | 0.530 | 0.598 |

**Table 7.** Robustness check of moderating analysis

| | (1)<br>High Digital | (2)<br>Low Digital | (3)<br>High OE | (4)<br>Low OE | (5)<br>High Ehold | (6)<br>Low Ehold | (7)<br>High IT_CEO | (8)<br>Low IT_CEO |
|---|---|---|---|---|---|---|---|---|
| HCAI | -0.0024* | -0.0008 | -0.0020 | -0.0042** | -0.0073*** | 0.0008 | -0.0024* | -0.0047** |
| | (0.0013) | (0.0030) | (0.0014) | (0.0019) | (0.0019) | (0.0015) | (0.0013) | (0.0022) |
| Size | -0.0018*** | -0.0006** | -0.0011*** | -0.0012*** | 0.0002 | -0.0022*** | -0.0008*** | -0.0017*** |
| | (0.0003) | (0.0003) | (0.0003) | (0.0003) | (0.0004) | (0.0003) | (0.0003) | (0.0003) |
| Roa | -0.0341*** | -0.0286*** | -0.0218*** | -0.0415*** | -0.0336*** | -0.0303*** | -0.0327*** | -0.0293*** |
| | (0.0050) | (0.0057) | (0.0055) | (0.0053) | (0.0053) | (0.0055) | (0.0048) | (0.0061) |
| Lev | -0.0236*** | -0.0194*** | -0.0262*** | -0.0181*** | -0.0237*** | -0.0197*** | -0.0223*** | -0.0200*** |
| | (0.0021) | (0.0022) | (0.0022) | (0.0021) | (0.0022) | (0.0022) | (0.0020) | (0.0024) |
| Age | -0.0051*** | -0.0042*** | -0.0038*** | -0.0057*** | -0.0057*** | -0.0035*** | -0.0052*** | -0.0038*** |
| | (0.0004) | (0.0004) | (0.0004) | (0.0004) | (0.0005) | (0.0004) | (0.0004) | (0.0004) |
| Capital | 0.0004* | 0.0010*** | 0.0004* | 0.0010*** | 0.0011*** | 0.0003* | 0.0006*** | 0.0009*** |
| | (0.0002) | (0.0002) | (0.0002) | (0.0002) | (0.0002) | (0.0002) | (0.0002) | (0.0002) |
| CashFlow | 0.0304*** | 0.0507*** | 0.0518*** | 0.0303*** | 0.0407*** | 0.0336*** | 0.0385*** | 0.0435*** |
| | (0.0085) | (0.0082) | (0.0083) | (0.0082) | (0.0084) | (0.0082) | (0.0079) | (0.0088) |
| BM | -0.0759*** | -0.0694*** | -0.0753*** | -0.0721*** | -0.0766*** | -0.0691*** | -0.0768*** | -0.0673*** |
| | (0.0020) | (0.0021) | (0.0021) | (0.0020) | (0.0021) | (0.0021) | (0.0019) | (0.0022) |
| Growth | 0.0130*** | 0.0127*** | 0.0122*** | 0.0135*** | 0.0137*** | 0.0116*** | 0.0123*** | 0.0137*** |
| | (0.0009) | (0.0008) | (0.0008) | (0.0008) | (0.0008) | (0.0008) | (0.0008) | (0.0008) |
| Bsize | -0.0031** | -0.0038*** | -0.0030** | -0.0036*** | -0.0037*** | -0.0027** | -0.0032*** | -0.0021 |
| | (0.0013) | (0.0013) | (0.0013) | (0.0013) | (0.0013) | (0.0013) | (0.0012) | (0.0014) |
| Top1 | -0.0093*** | -0.0045** | -0.0076*** | -0.0052*** | -0.0038* | -0.0073*** | -0.0057*** | -0.0069*** |
| | (0.0019) | (0.0019) | (0.0019) | (0.0020) | (0.0020) | (0.0019) | (0.0018) | (0.0021) |
| INST | -0.0001 | -0.0021* | -0.0005 | -0.0022* | -0.0028** | 0.0012 | -0.0013 | -0.0017 |
| | (0.0013) | (0.0012) | (0.0013) | (0.0012) | (0.0014) | (0.0018) | (0.0012) | (0.0013) |
| Constant | 0.1792*** | 0.1436*** | 0.1596*** | 0.1606*** | 0.1333*** | 0.1771*** | 0.1557*** | 0.1630*** |
| | (0.0060) | (0.0063) | (0.0059) | (0.0065) | (0.0073) | (0.0056) | (0.0060) | (0.0063) |
| Observations | 8356 | 8352 | 8354 | 8355 | 8356 | 8354 | 9318 | 7393 |
| R-squared | 0.4132 | 0.4317 | 0.4242 | 0.4120 | 0.4004 | 0.4175 | 0.4041 | 0.4248 |

Standard errors in parentheses. *p < 0.1, **p < 0.05, ***p < 0.01